\documentclass[12pt]{article}

\usepackage{graphicx}
\usepackage{amsmath,amssymb,amsfonts}
\usepackage{algpseudocode,algorithm}
\usepackage{subcaption}
\usepackage{makecell}

\usepackage{times}
\usepackage{color}
\usepackage{multirow}
\usepackage[authoryear]{natbib}
\usepackage{rotating}
\usepackage{bbm}
\usepackage{latexsym}
\usepackage{url}

\newcommand{\captionfonts}{\normalsize}

\makeatletter
\long\def\@makecaption#1#2{%
  \vskip\abovecaptionskip
  \sbox\@tempboxa{{\captionfonts #1: #2}}%
  \ifdim \wd\@tempboxa >\hsize
    {\captionfonts #1: #2\par}
  \else
    \hbox to\hsize{\hfil\box\@tempboxa\hfil}%
  \fi
  \vskip\belowcaptionskip}
\makeatother

\begin{document}
\hspace{13.9cm}

\ \vspace{20mm}\\
{\LARGE DROP: Distributional and Regular Optimism and Pessimism for Reinforcement Learning}

\ \\
{\bf \large Taisuke Kobayashi$^{\displaystyle 1}$}\\
{$^{\displaystyle 1}$National Institute of Informatics (NII) and The Graduate University for Advanced Studies (SOKENDAI),
        2-1-2 Hitotsubashi, Chiyoda-ku, Tokyo, 101-8430, Japan.}\\
%


\thispagestyle{empty}
\markboth{}{NC instructions}
\ \vspace{-0mm}\\
\begin{center} {\bf Abstract} \end{center}
In reinforcement learning (RL), temporal difference (TD) error is known to be related to the firing rate of dopamine neurons.
It has been observed that each dopamine neuron does not behave uniformly, but each responds to the TD error in an optimistic or pessimistic manner, interpreted as a kind of distributional RL.
To explain such a biological data, a heuristic model has also been introduced with learning rates asymmetric for the positive and negative TD errors.
However, this heuristic model is not theoretically-grounded and unknown whether it can work as a RL algorithm.
This paper therefore introduces a novel theoretically-grounded model with optimism and pessimism, which is derived from control as inference.
In combination with ensemble learning, a distributional value function as a critic is estimated from regularly introduced optimism and pessimism.
Based on its central value, a policy in an actor is improved.
This proposed algorithm, so-called DROP (distributional and regular optimism and pessimism), is compared on dynamic tasks.
Although the heuristic model showed poor learning performance, DROP demonstrated excellent performance in all tasks with high generality.
In addition, DROP achieved learning performance comparable to the state-of-the-art algorithms.
In other words, it was suggested that DROP is a new model that can elicit the potential contributions of optimism and pessimism.

\section{Introduction}

Among the machine learning technologies that have made dramatic progress in recent years, reinforcement learning (RL) \citep{sutton2018reinforcement} stands out for its uniqueness (e.g. online data collection by agents and indirect optimization through rewards).
RL, which was developed and established in the context of optimal control, has begun to be used in a wide range of fields:
such as robotics \citep{wu2023daydreamer};
game AI \citep{oh2021creating};
and autonomous driving \citep{cui2021autonomous}.

On the other hand, RL deals with decision-making problems that consider the agent's future, and analogies with those of animals (including human) have been pointed out.
Based on that, various research groups have found several correspondences between the decision-making data of animals and RL or its components.
For example, the temporal difference (TD) error in RL is proportional to the firing rate of dopamine neurons \citep{schultz1993responses,starkweather2021dopamine}.
The discount factor in RL, i.e. how far into the future to be considered, is related to the amount of serotonin \citep{tanaka2007serotonin,koolschijn2024resources}.
The intrinsic motivation manipulates the external rewards in RL, making human adapt to various situations \citep{blain2021intrinsic,molinaro2023intrinsic}.
The perceptual and value-based decision making is in the duality according to control as inference \citep{levine2018reinforcement}, and it might be implemented on the sensory and motor cortical circuits \citep{doya2021canonical}.
Thus, designing new RL algorithms can help us understand the decision-making models of animals.
Conversely, findings from animal data can provide hints for designing new RL algorithms.

\begin{figure}[tb]
    \centering
    \includegraphics[keepaspectratio=true,width=0.96\linewidth]{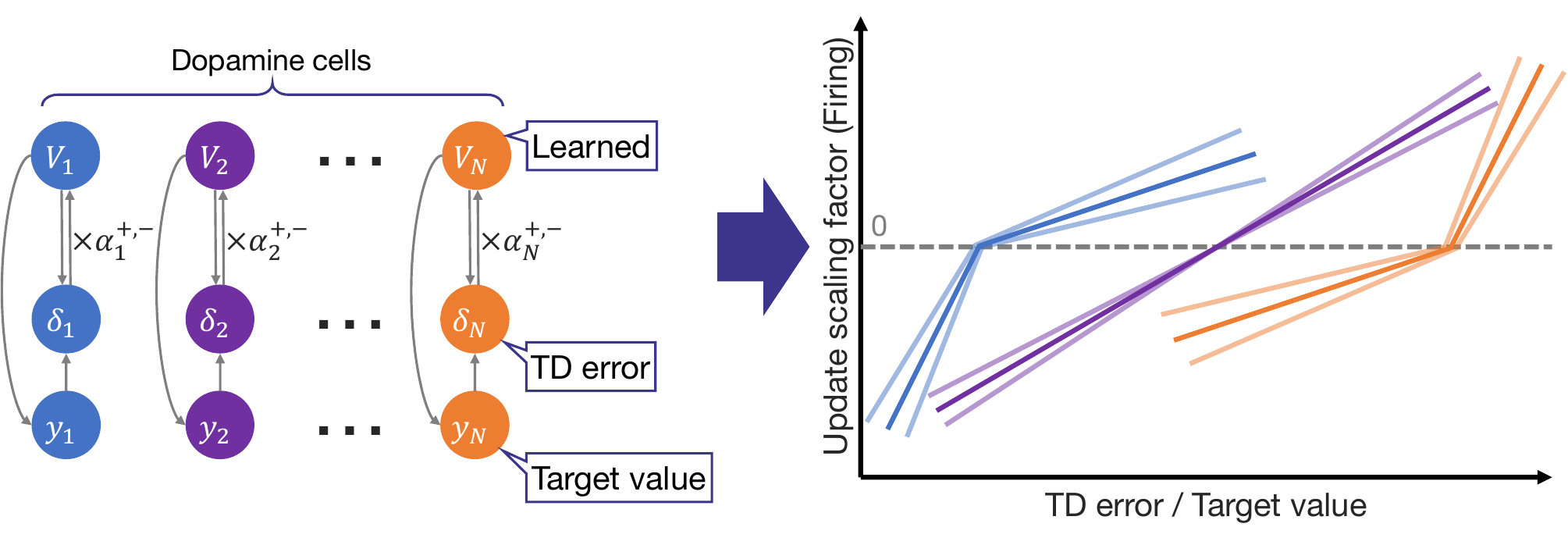}
    \caption{Heuristic model designed in the previous studies \citep{dabney2020distributional,muller2024distributional}:
    because the model is an ensemble model with each dopamine neuron estimating its value function individually, the origin at which the TD error becomes zero is different for each neuron;
    each neuron has an asymmetric learning rate depending on the sign of TD error, so that some neurons learn optimistically, preferring a better outcome than predicted, while others are pessimistic for preferring a worse outcome than predicted.
    }
    \label{fig:background}
\end{figure}

Among them, this paper focuses on the latest studies on TD errors and the firing of dopamine neurons \citep{dabney2020distributional,muller2024distributional}.
In these papers, it is reported that dopamine neurons fire at different times.
This suggests that each dopamine neuron independently learns its own value function that minimizes the corresponding TD error, and that decisions are made by combining these value functions.
Furthermore, some dopamine neurons have asymmetrically biased firing patterns depending on the positive or negative TD error, leading to the distributional estimation of the value function as the distributional RL \citep{bellemare2017distributional,tano2020local}.
To represent these characteristics, a heuristic model was introduced in which the learning rate $\alpha$ is asymmetrically given for the positive and negative TD errors, as shown in Fig.~\ref{fig:background}.

The above heuristic model seems to be useful enough as an RL algorithm.
In fact, this is one option (i.e. expectile regression) assuming that the value function is expanded to a distribution model, which encompasses not only the average return but also the worst-case and best-case scenarios that could occur stochastically \citep{rowland2019statistics}.
Intuitively, however, there is a concern that it may destabilize RL because of the unsmoothness in the update amounts based on TD errors.
In addition, learning of policies employs the sample-averaged gradient naively, without directly introducing such asymmetry \citep{nam2021gmac}.
We should derive and verify a model with optimism and pessimism that is more theoretically plausible as a RL algorithm for learning both value and policy functions.
That is, by finding optimism and pessimism in a bottom-up manner from the learning theory of RL, it can be expected that distributional RL will function more appropriately, and such a flow might be more natural considering the process of biological evolution.

To this end, this paper focuses on the previous work \citep{kobayashi2022optimistic}.
In this paper, the learning rule for value and policy functions in RL is rederived based on the concept of control as inference \citep{levine2018reinforcement}.
In other words, the problem setting of RL is changed from maximizing the cumulative rewards in the future to optimizing the probability of optimality defined by them, and the approximated gradients for the new problem setting are shown to be consistent with the conventional learning rule.
By revisiting the arbitrary definitions behind this problem setting, a theoretically-grounded optimistic learning rule of both value and policy functions can also be derived with its parameterization, enabling as an RL algorithm to accelerate exploration.

Based on this previous work, the first objective of this paper is to find a new derivation method, which converts the optimism into pessimism in a similar way.
For this purpose, the original definition of optimality used in the derivation of optimism is modified to its inversion.
Afterward, using the same procedure of computing the approximated gradients as in the previous study, the pessimistic learning rule of both value and policy functions is parameterized as well.
It is also pointed out that these optimism and pessimism can be expressed and adjusted with a unified parameter by integrating their learning rules.

Next, a RL algorithm is developed to learn all components of an ensemble model for representing the value distribution with different optimistic and pessimistic parameters.
For this purpose, it would be efficient to estimate values by placing each component so that the cumulative probability of the value distribution falls at regular intervals as much as possible.
However, since the unified parameter is defined in real space, designing to meet such regular spacing is difficult.
Based on the boundedness of the nonlinear (i.e. optimistic or pessimistic) TD errors, the original parameter is invertibly transformed to bounded one, which can intuitively be regularly spaced on its bounded space with reasonable approximation.
In addition, by interpreting the obtained value functions as a multi-objective optimization problem, policy improvement is implemented based on the median of nonlinear TD errors.

The developed RL algorithm, so-called DROP (distributional and regular optimism and pessimism), is compared with the heuristic model in the previous studies \citep{dabney2020distributional,muller2024distributional} and benchmark performance using two dynamics simulators \citep{todorov2012mujoco,coumans2016pybullet}.
The results show that the heuristic model destabilizes learning by failing to properly converge the TD error of the pessimistic value function.
In contrast, DROP succeeds in learning efficiently and stably in all benchmark tasks to a level where the tasks could qualitatively be considered successful.
Additionally, DROP demonstrates the better learning performance to the state-of-the-art (non-distributional and distributional) algorithms \citep{haarnoja2018soft,duan2025distributional} in \citep{tunyasuvunakool2020dm_control}.

\subsection{Related work}

Four concepts in RL relevant to this paper are introduced here to clarify the technological position of the proposed method.

\subsubsection{Bio-inspired RL}

As explained, RL has high affinity with decision-making of animals.
Inspired by them, various algorithms and designs have been proposed, contributing to the improvement of RL performance.
For example, the biological Levy walk has been conceptually introduced into the exploration strategy in RL, improving sample efficiency of optimization \citep{kobayashi2019student,wang2024bioinspired}.
As a means of encouraging more active exploration, various bonuses to reward have been designed based on the concept of intrinsic motivation \citep{devidze2022exploration,karbasi2024embodied}.
Since animals have different information processing is conducted in reward and punishment, the extended RL framework, which separately deals with reward and punishment, has been designed, improving the performance of safely finding the optimal solution \citep{kobayashi2019reward,wang2021modular}.
The world model, which subjectively represents the state in RL according to the agent's observation, allows the agent to simulate the interaction with environment, facilitating learning and/or avoiding risky behaviors in real world \citep{ha2018world,wu2023daydreamer}.

As explained in the introduction, this study reflects the fact that dopamine neurons fire differently in individuals and are biased in an optimistic or pessimistic direction.
In other words, one can expect that developing an optimistic and/or pessimistic RL algorithm, like that of animals, will not only facilitate our understanding of decision-making in animals, but will also elicit the benefits of optimism and pessimism, as described below.

\subsubsection{Optimistic RL}

Optimism in RL refers to a high estimate of the sum of future rewards (i.e. the value function).
Optimism can be incorporated into efficient exploration methods because it encourages the acquisition of diverse experiences without fear of the unknown states.
The simplest and classic method is to make the inital value function larger \citep{strehl2006pac}.
This concept is extended to the recent deep RL framework by considering whether the encounter situation is unexplored \citep{lobel2022optimistic}.
It is also a kind of optimism to introduce an upper confidence bound (UCB), which selects more potentially valuable actions by taking into account the uncertainty in the value function \citep{chen2017ucb,ciosek2019better}.
UCB is sometimes added to the learning of the value functionn as an exploration bonus \citep{bai2021principled,kobayashi2023reward}.

The optimism of this study biases the amount of updating against the TD error, which can be considered similar to that of UCB, but based on the TD error rather than on uncertainty.
While the cost of estimating uncertainty is unnecessary, the bias, a kind of reward shaping \citep{ng1999policy}, depends not only on the state but also on the action, and thus the value function is estimated to be higher than the expectation.

\subsubsection{Pessimistic RL}

Contrary to optimism, pessimism, as the name implies, lowers the value for decision-making.
This leads to conservative learning and is expected to reduce the risk of falling into a local solution \citep{haarnoja2018soft,fujimoto2018addressing}.
Pessimism is particularly useful in offline RL, where out-of-distribution actions are problematic, because it encourages the user to stay in a known state space \citep{shi2022pessimistic,bai2022pessimistic}.
It is also more sensitive to risk, making it easier to consider worst-case scenarios and ensure safety \citep{wachi2020safe,schneider2024learning}.

The pessimism of this study underestimates the value function, expecting conservative and safe learning as well as the above related studies.
Note, however, that it does not take pessimism into account as markedly as in related studies, but only to the extent that it is biased.

As suggested by their respective characteristics, optimism and pessimism are opposing concepts.
Therefore, each has its own advantages and disadvantages.
To bring out these advantages only, the complementary combination of these two has been studied actively \citep{curi2021combining,moskovitz2021tactical,bura2022dope,nauman2025decoupled}.

\subsubsection{Distributional RL}

Distributional RL handles the randomness of the return due to stochastic state transitions.
It predicts the return distributionally, which encompasses both better-case and worse-case return estimates;
and for example, by setting the goal of maximizing the return biased from the distribution center, this would lead to the optimistic policies (expecting fortunate transitions with an ideal outcome) or pessimistic ones (preparing for a troublesome scenario).

There are two ways to learn such a value distribution: explicitly assuming a distribution model or implicitly learning it.
While the former (e.g. \citep{bellemare2017distributional,duan2025distributional}) is simple to implement, it requires (ad-hoc) handling of learning problems caused by modeling errors.
In the latter, algorithms using quantile regression are particularly noteworthy \citep{dabney2018distributional,rowland2024analysis}.
Although this theoretically minimizes Wasserstein distance, various algorithms have been developed to close the gap with practical implementation \citep{dabney2018implicit,luo2022distributional}.
Expectile regression has also been used instead of the quantile regression \citep{rowland2019statistics,tano2020local}, which corresponds to the RL with asymmetric learning rate introduced in the introduction.
While easy to implement, there are some problems such as difficulty in sampling, so methods using both quantile and expectile has recently been proposed \citep{jullien2023distributional}.

In both cases, the learning rule is expressed in a piecewise linear manner, such as with the asymmetric learning rate, but in the proposed method, a (smooth) nonlinear transformation of the TD error corresponds to the update amount.
In other words, while the proposed method should represent the value distribution in the similar way as the conventional methods, it might represent neither quantile nor expectile as a statistic.
This paper focuses on the learning performance (in particular, generalization performance to tasks) of the smooth nonlinearity derived from the proposed method, rather than on clarifying the nature of the statistic implicitly learned by the proposed method.

The main focus of these studies is on learning distributional value functions, and parallel work has been done on policy improvements according to the resulting distributional value functions.
The baseline actor-critic algorithm is extended to accommodate the introduction of a distributional value function, but basically a point estimate of the distributional value distribution is employed in the usual policy gradient method \citep{nam2021gmac,schneider2024learning,duan2025distributional}.
This approach is followed in this paper as well, using a point estimate of the median error of the ensemble of values.

\section{Preliminaries}

\subsection{Reinforcement learning}

RL aims to optimize a policy of a learnable agent so that the accumulation of future rewards from an unknown environment (so-called return) is maximized \citep{sutton2018reinforcement}.
This problem assumes Markov decision process (MDP) with the tuple $(\mathcal{S}, \mathcal{A}, \mathcal{R}, p_0, p_e)$.
Specifically, the state $s \in \mathcal{S}$ is sampled from the environment with either of probabilities, $s \sim p_0(s)$ as the initial random state or $s \gets s^\prime \sim p_e(s^\prime \mid s, a)$ as the state transition.
The agent decides the action $a \in \mathcal{A}$ according to $s$, using the policy $a \sim \pi(a \mid s)$\footnote{$a$ is actually sampled from a baseline policy $b(a \mid s)$ to formally distinguish the policy at sampling time from $\pi(a \mid s)$ to be optimized.}, which can be learned by updating its parameters $\phi$.
$a$ yields the state transition $s^\prime \sim p_e(s^\prime \mid s, a)$, and then, this one step is evaluated as the reward $r = f_r(s, a) \in \mathcal{R} \subseteq \mathbb{R}$.

By sequentially repeating the above step with the experience data $(s, a, s^\prime, r)$, the agent obtains the following return $R_t$ from the current time step $t$.
\begin{align}
    R_t = \sum_{k=0}^\infty \gamma^k r_{t+k}
    \label{eq:return}
\end{align}
where $\gamma \in [0, 1)$ denotes the discount factor.

Under this MDP, $\pi(a \mid s)$ is optimized as follows:
\begin{align}
    \pi^*(\cdot \mid s) = \arg \max_{\pi} \mathbb{E}_{p_e, \pi}[R_t \mid s_t = s]
    \label{eq:prob_rl}
\end{align}
Note that $\mathbb{E}_{p_e, \pi}[R_t \mid s_t = s]$ is defined as the (state) value function $V^\pi(s)$, which is approximated with its parameters $\theta$.
For the brevity, $V^\pi$ is described $V$ later.

\subsection{Optimism from control as inference}

In the previous work, a theoretically-grounded optimistic RL has been derived \citep{kobayashi2022optimistic}.
Following it, this paper introduces a stochastic variable for optimality, $O = \{0, 1\}$, which is relevant to the return.
The conditinal probability of $O=1$ (i.e. the future is optimal) is modeled with $\beta \in \mathbb{R}_+$ the inverse temperature parameter.
\begin{align}
    p(O=1 \mid s) &= e^{\beta (V(s) - \overline{R})}
    \label{eq:pv_optim} \\
    p(O=1 \mid s, a) &= e^{\beta (Q(s,a) - \overline{R})}
    \label{eq:pq_optim}
\end{align}
where $Q(s, a) = \mathbb{E}_{p_e, \pi}[R_t \mid s_t = s, a_t = a]$ denotes the action value function and can be approximated to be $r + \gamma V(s^\prime)$ according to Bellman equation and Monte Carlo approximation.
$\overline{R}$ is the implicit maximum return to satisfy $p(O=1) \in (0, 1)$.
This is necessary to satisfy the definition of probability, but does not require a specific value since it will be excluded in later calculations.
Note that the probability of $O=0$ can be easily given as $O$ is binary.

Here, the optimal and non-optimal policies can be theoretically inferred (although they cannot be directly used for generating $a$).
Specifically, with the baseline policy $b(a \mid s)$ for sampling actions (a.k.a. the old $\pi$), $\pi(a \mid s, O)$ is given according to Bayesian theorem as follows:
\begin{align}
    &\pi(a \mid s, O)
    = \cfrac{p(O \mid s, a) b(a \mid s)}{p(O \mid s)}
    \nonumber \\
    &= \begin{cases}
        \cfrac{e^{\beta (Q(s,a) - \overline{R})}}{e^{\beta (V(s) - \overline{R})}} b(a \mid s) =: \pi^+(a \mid s) & O = 1
        \\
        \cfrac{1 - e^{\beta (Q(s,a) - \overline{R})}}{1 - e^{\beta (V(s) - \overline{R})}} b(a \mid s) =: \pi^-(a \mid s)& O = 0
    \end{cases}
    \label{eq:policy_bayes_optim}
\end{align}

Under the above setup, the following two optimization problems with Kullback-Leibler (KL) divergences are solved.
\begin{align}
    &\min_\theta \mathbb{E}_{p_e, b}[\mathrm{KL}(p(O \mid s, a) \mid p(O \mid s))]
    \label{eq:prob_val} \\
    &\min_\phi \mathbb{E}_{p_e}[\mathrm{KL}(\pi^+(a \mid s) \mid \pi(a \mid s)) - \mathrm{KL}(\pi^-(a \mid s) \mid \pi(a \mid s))]
    \label{eq:prob_pol}
\end{align}
where $\mathrm{KL}(p_1 \mid p_2) = \mathbb{E}_{x \sim p_1}[\ln p_1(x) - \ln p_2(x)]$.
Note again that $\theta$ and $\phi$ denote the parameters for $V$ and $\pi$, respectively.
The first minimization problem aims to get closer to $p(O \mid s, a)$ with more reliable information about optimality than $p(O \mid s)$ with $\theta$.
The second intends to make $\pi$ with $\phi$ closer to the optimal policy, but away from the non-optimal one.
Note that as $b$ is the old $\pi$, it is independent of $\phi$ for the current $\pi$.

The approximated gradients for solving these are derived below (for detailed derivation processes, see the literature \citep{kobayashi2022optimistic} and the derivation process for the pessimistic RL in the next section).
\begin{align}
    g_\theta^\mathrm{optim} &= \mathbb{E}_{p_e, b}[-\beta^{-1}(e^{\beta(r + \gamma V(s^\prime) - V(s))} - 1) \nabla_\theta V(s)]
    \label{eq:grad_val_optim} \\
    g_\phi^\mathrm{optim} &= \mathbb{E}_{p_e, b}[-\beta^{-1}(e^{\beta(r + \gamma V(s^\prime) - V(s))} - 1) \nabla_\phi \ln \pi(a \mid s)]
    \label{eq:grad_pol_optim}
\end{align}
As $Q(s,a) - V(s) \simeq r + \gamma V(s^\prime) - V(s)$ is also known to be the TD error, $\delta$, both gradients are nonlinearly related to $\delta$ as the nonlinear TD error, $f(\delta) := \beta^{-1}(e^{\beta \delta} - 1)$.
Due to the positive $\beta$, $d f(\delta) / d \delta \geq 0$ and $d^2 f(\delta) / d \delta^2 \geq 0$ hold: i.e. $f(\delta)$ is the monotonically increasing convex function w.r.t. $\delta$.
In other words, the update amount of $\delta > 0$ is larger than that of $\delta < 0$, making RL optimistic.

\section{Derivation of pessimism}

\subsection{Inversion of optimality definition}

\begin{figure}[tb]
    \centering
    \includegraphics[keepaspectratio=true,width=0.84\linewidth]{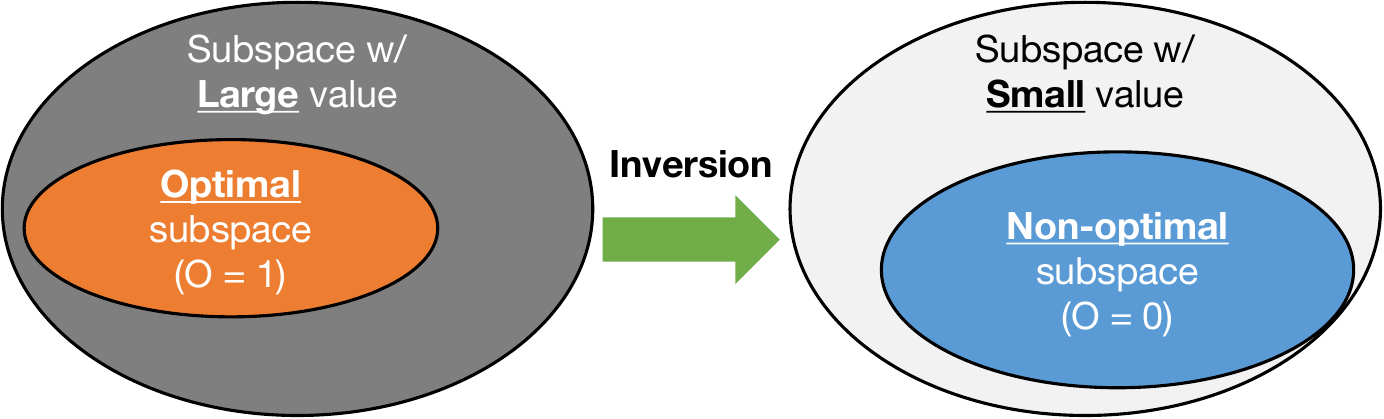}
    \caption{Inversion of the definition of optimality:
    the definition of optimality in the previous studies \citep{levine2018reinforcement,kobayashi2022optimistic} was that the larger the value function, the more optimal (i.e. $O=1$) it is, as shown on the left side;
    the inversion is similar but different that the smaller the value function, the more non-optimal (i.e. $O=0$) it is, as shown on the right side.
    }
    \label{fig:inversion}
\end{figure}

The derivation of the above optimistic RL starts from the definition of $p(O)$.
Remarkably, the exponential function in it is carried over to the approximated gradients, which yield the optimism.
With this in mind, this paper rethinks the definition of $p(O)$ in order to theoretically formulate the pessimism.

We have to notice that this definition can be assumed arbitrarily, except that $p(O=1)$ is monotonically increasing with respect to the return.
Among the several possible definitions, this paper takes the following definition as a starting point, expecting that the pessimism, which is the inversion of the optimism, would emerge from the inversion of the definition in which the optimism emerged (see Fig.~\ref{fig:inversion}).
\begin{align}
    p(O=0 \mid s) &= e^{\beta (-V(s) + \underline{R})}
    \label{eq:pv_pessim}\\
    p(O=0 \mid s, a) &= e^{\beta (-Q(s, a) + \underline{R})}
    \label{eq:pq_pessim}
\end{align}
where $\beta \in \mathbb{R}_+$ denotes the inverse temperature parameter as well as the original definition.
$\underline{R}$ is the implicit minimum return to satisfy $p(O=0) \in (0, 1)$.
That is, as the return increases, the probability of $O=0$ (i.e. the future is non-optimal) monotonically decreases, and simulutaneously, the probability of $O=1$ monotonically increases.

With this definition, the optimal and non-optimal policies are again inferred as follows:
\begin{align}
    \begin{split}
        \pi^+(a \mid s) &= \cfrac{1 - e^{\beta (-Q(s, a) + \underline{R})}}{1 - e^{\beta (-V(s) + \underline{R})}} b(a \mid s)
        \\
        \pi^-(a \mid s) &= \cfrac{e^{\beta (-Q(s, a) + \underline{R})}}{e^{\beta (-V(s) + \underline{R})}} b(a \mid s)
    \end{split}
    \label{eq:policy_bayes_pessim}
\end{align}

\subsection{Derivation of pessimism}

Under the new definition above, let's consider solving eqs.~\eqref{eq:prob_val} and ~\eqref{eq:prob_pol}.
The almost same procedure as in the case of the above optimistic RL is conducted to derive the approximated gradients.

Specifically, $g_\theta^{\mathrm{pessim}}$, the gradient for $\theta$ in the value function $V$, is first derived as follows:
\begin{align}
    g_\theta^{\mathrm{pessim}} &=
    - p(O=1 \mid s, a) \nabla_\theta \ln p(O=1 \mid s)
    \nonumber \\
    &\quad - p(O=0 \mid s, a) \nabla_\theta \ln p(O=0 \mid s)
    \nonumber \\
    &= - (1 - e^{\beta (-Q(s,a) + \underline{R})}) \cfrac{e^{\beta (-V(s) + \underline{R})} \beta \nabla_\theta V(s)}{1 - e^{\beta (-V(s) + \underline{R})}}
    \nonumber \\
    &\quad + e^{\beta (-Q(s,a) + \underline{R})} \beta \nabla_\theta V(s)
    \nonumber \\
    &= \beta \nabla_\theta V(s) \cfrac{e^{\beta (-Q(s,a) + \underline{R})} - e^{\beta (-V(s) + \underline{R})}}{1 - e^{\beta (-V(s) + \underline{R})}}
    \nonumber \\
    &\propto \nabla_\theta V(s) \beta^{-1} \left( \cfrac{e^{\beta (-Q(s,a) + \underline{R})}}{e^{\beta (-V(s) + \underline{R})}} - 1 \right)
    \nonumber \\
    &\simeq \nabla_\theta V(s) \beta^{-1} (e^{-\beta (r + \gamma V(s^\prime) - V(s))} - 1)
    \label{eq:grad_val_pessim_0}
\end{align}
where the proportion is obtained by dividing the third line by $\beta^2 e^{\beta (-V(s) + \underline{R})} (1 - e^{\beta (-V(s) + \underline{R})})^{-1} > 0$, and $Q(s, a)$ is finally approximated as $r + \gamma V(s^\prime)$.
Note that for simpicity, only the inside of the expectation operation $\mathbb{E}_{p_e, b}[\cdot]$ is considered here.

Similarly, $g_\phi^{\mathrm{pessim}}$, the gradient for $\phi$ in the policy $\pi$, is approximately derived.
Again, only the inside of the expectation operation $\mathbb{E}_{p_e}[\cdot]$ is considered for simplicity.
\begin{align}
    g_\phi^{\mathrm{pessim}} &=
    \mathbb{E}_{b} \Biggl[
    - \cfrac{1 - e^{\beta (-Q(s, a) + \underline{R})}}{1 - e^{\beta (-V(s) + \underline{R})}} \nabla_\phi \ln \pi(a \mid s)
    \nonumber \\
    &\quad\quad\quad\quad\quad\quad + \cfrac{e^{\beta (-Q(s, a) + \underline{R})}}{e^{\beta (-V(s) + \underline{R})}} \nabla_\phi \ln \pi(a \mid s)
    \Biggr]
    \nonumber \\
    &= \mathbb{E}_{b} \Biggl[
    \cfrac{\nabla_\phi \ln \pi(a \mid s)}{e^{\beta (-V(s) + \underline{R})}} \cfrac{e^{\beta (-Q(s, a) + \underline{R})} - e^{\beta (-V(s) + \underline{R})}}{1 - e^{\beta (-V(s) + \underline{R})}}
    \Biggr]
    \nonumber \\
    &\propto \mathbb{E}_{b} \Biggl[
    \nabla_\phi \ln \pi(a \mid s) \beta^{-1} \left( \cfrac{e^{\beta (-Q(s,a) + \underline{R})}}{e^{\beta (-V(s) + \underline{R})}} - 1 \right)
    \Biggr]
    \nonumber \\
    &\simeq \mathbb{E}_{b} \Biggl[ \nabla_\theta V(s) \beta^{-1} (e^{-\beta (r + \gamma V(s^\prime) - V(s))} - 1) \Biggr]
    \label{eq:grad_pol_pessim_0}
\end{align}
where the proportion is given by dividing the second line by $\beta (1 - e^{\beta (-V(s) + \underline{R})})^{-1} > 0$.

The two gradients obtained are summarized below.
\begin{align}
    g_\theta^\mathrm{pessim} &= \mathbb{E}_{p_e, b}[-\beta^{-1}(1 - e^{-\beta(r + \gamma V(s^\prime) - V(s))}) \nabla_\theta V(s)]
    \label{eq:grad_val_pessim} \\
    g_\phi^\mathrm{pessim} &= \mathbb{E}_{p_e, b}[-\beta^{-1}(1 - e^{-\beta(r + \gamma V(s^\prime) - V(s))}) \nabla_\phi \ln \pi(a \mid s)]
    \label{eq:grad_pol_pessim}
\end{align}
With $f(\delta) := \beta^{-1}(1 - e^{-\beta \delta})$, $d f(\delta) / d \delta \geq 0$ and $d^2 f(\delta) / d \delta^2 \leq 0$ hold: i.e. $f(\delta)$ is the monotonically increasing concave function w.r.t. $\delta$.
In other words, the update amount of $\delta > 0$ is smaller than that of $\delta < 0$, making RL pessimistic.

\begin{figure}[tb]
    \centering
    \includegraphics[keepaspectratio=true,width=0.84\linewidth]{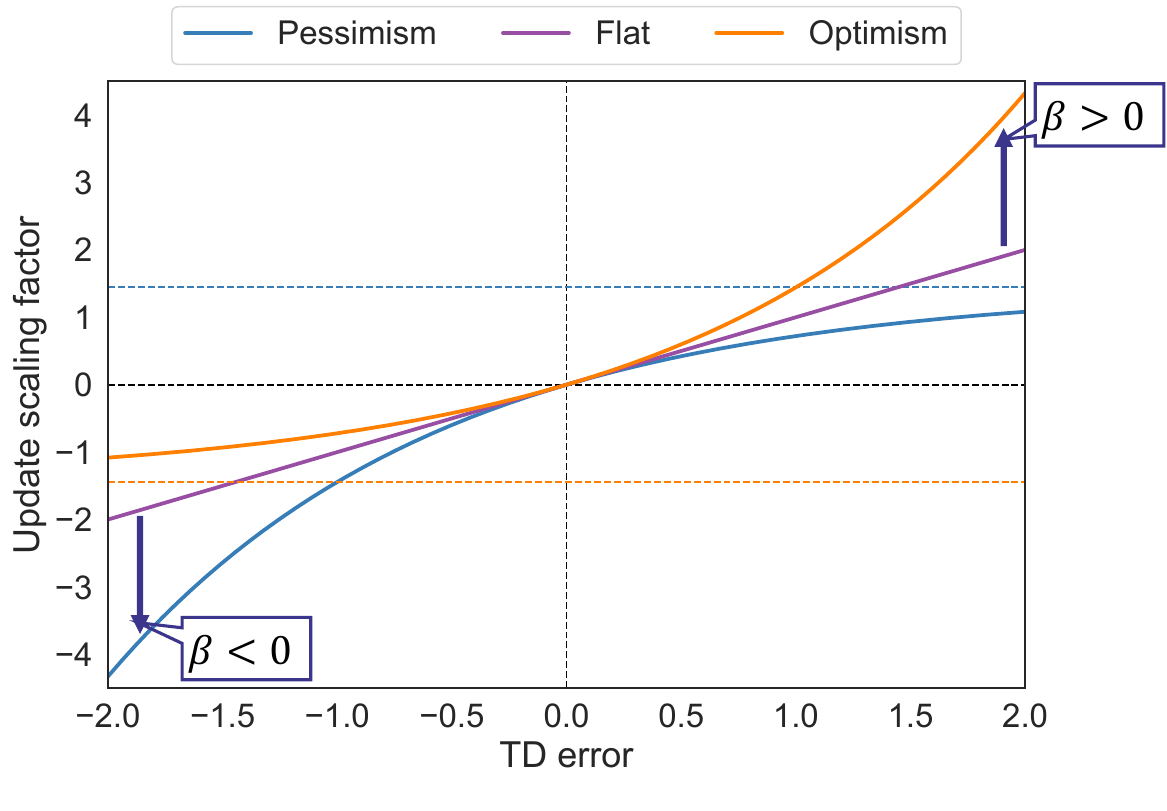}
    \caption{Optimistic and pessimistic TD errors parameterized by $\beta \in \mathbb{R}$:
    when $\beta > 0$, the update scale is positively biased with optimism from the original TD error;
    symmetrically, pessimism can be obtained with $\beta < 0$.
    }
    \label{fig:op_tderr}
\end{figure}

Compared to eqs.~\eqref{eq:grad_val_optim} and~\eqref{eq:grad_pol_optim}, these two equations look similar.
In fact, without loss of generality, if a minus sign is included into $\beta$ (i.e. $\beta < 0$ is assumed), they can be expressed in exactly the same way.
That is, the optimistic and pessimistic update rules can be adapted by $\beta \in \mathbb{R}$, as defined in the following equation and illustrated in Fig.~\ref{fig:op_tderr}.
\begin{align}
    f_\beta(\delta) &:= \begin{cases}
        \delta & \beta = 0
        \\
        \beta^{-1}(e^{\beta \delta} - 1) & \beta \neq 0
    \end{cases}
    \label{eq:drop} \\
    g_\theta &= \mathbb{E}_{p_e, b}[-f_\beta(\delta) \nabla_\theta V(s)]
    \label{eq:grad_val} \\
    g_\phi &= \mathbb{E}_{p_e, b}[-f_\beta(\delta) \nabla_\phi \ln \pi(a \mid s)]
    \label{eq:grad_pol}
\end{align}
Note that the limits $\beta$ to $0$ from both positive and negative sides converge to $\beta^{-1}(e^{\beta \delta} - 1) \to \delta$.
Therefore, $f_{\beta=0}(\delta) = \delta$ is given.
In addition, as can be easily seen in the figure, when $\delta=0$, $f_\beta(0) = 0$ and $d f_\beta(0) / d \delta = 1$ hold for any $\beta$, yielding $f_{\beta_1}(\delta) \geq f_{\beta_2}(\delta)$ with $\beta_1 > \beta_2$.
In other words, $f_\beta(\delta) - \delta \geq 0$ holds for $\beta > 0$, and $f_\beta(\delta) - \delta \leq 0$ for $\beta < 0$.

\section{Implementation of DROP}

\subsection{Distributional value function with ensemble model}

As suggested in the literature \citep{dabney2020distributional}, the optimistic and pessmistic update rules for the value function would achieve the ones biased positively and negatively, respectively.
By combining the value functions learned with the optimistic-to-pessimistic update rules, the distributional value function can be modeled implicitly, although errors might occur compared to learning an explicit well-modeled distribution \citep{tano2020local}.
However, this might improve the accuracy of value estimation compared to the standard point estimate, and also allows for applications such as a risk-sensitive optimization like \citet{kim2022trc}, although they are out of the scope in this paper.

\begin{figure}[tb]
    \centering
    \includegraphics[keepaspectratio=true,width=0.96\linewidth]{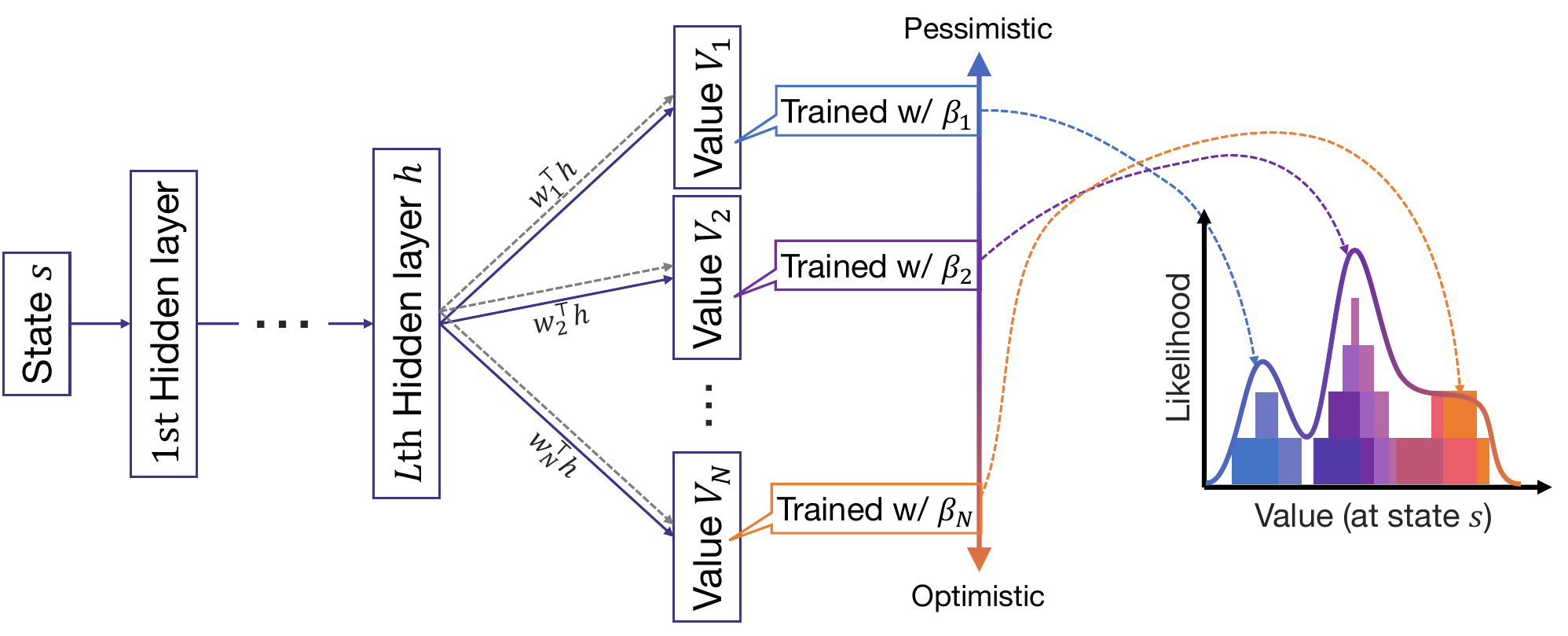}
    \caption{Distributional value function modeled by an ensemble model of multiple value functions with different optimism/pessimism:
    the network parameters before the output layer are all shared between $N$ value functions;
    with randomly fixed weights for diversity, the output layer separately estimates the respective value functions, which are trained with the corresponding $\beta_1, \ldots, \beta_N$;
    as value functions trained with biased TD errors are biased as well, their ensemble can represent the distribution of value functions.
    }
    \label{fig:value_dist}
\end{figure}

Specifically, eq.~\eqref{eq:grad_val} is extended to the distributional version simply with an ensemble model of $N$ estimates of $V$ with the respective parameters $\boldsymbol{\theta} = [\theta_1, \theta_2, \ldots, \theta_N]$.
These are separately optimized with the different inverse temperature parameters, $\boldsymbol{\beta} = [\beta_1, \beta_2, \ldots, \beta_N]$, as follows:
\begin{align}
    \theta_i &\gets \theta_i - \alpha \mathbb{E}_{(s, a, s^\prime, r) \sim \mathcal{D}}[-f_{\beta_i}(\delta_i) \nabla_{\theta_i} V_i(s)]
    \label{eq:value_dist} \\
    \delta_i &= r + \gamma \bar{V}_i(s^\prime) - V_i(s)
    \nonumber
\end{align}
where $\alpha \in \mathbb{R}_+$ denotes the learning rate and $\mathcal{D}$ is the replay buffer storing the experience data in a FIFO manner.
$\bar{V}_i$ denotes the value functions computed with target networks with $\bar{\theta}_i$ for stabilizing learning~\citep{kobayashi2024consolidated}.
As a remark, this ensemble model is implemented with a shared model before outputs and $N$ separated linear mappings for reducing its computational complexity (see Fig.~\ref{fig:value_dist}).
Note that although the previous studies \citep{dabney2020distributional,muller2024distributional} sample one value function for $s^\prime$ to compute all the TD errors, this paper employs the local version, which computes the above update in each cell independently, according to biological feasibility pointed out in the literature \citep{tano2020local}.
Although it has also been pointed out that independent ensemble value updates cannot represent value distributions, this implementation might expect us to mitigate this issue by sharing the features used to calculate values, as described above (also see Appendix~\ref{app:distribution}).

Next, $\boldsymbol{\beta}$ is designed to estimate the values with regular intervals in terms of the cumulative probability as much as possible.
Unfortunately, $\beta_i \in \mathbb{R}$ makes their balanced placement difficult.
The effective range of $\beta_i$ would be implicitly affected by the scale of the TD error.

To resolve these issues, the upper or lower bound of $f(\delta)$ is utilized based on the previous work \citep{kobayashi2022optimistic}: with $\eta, \beta > 0$, the lower bound is given as $-\beta^{-1}$; and with $\eta, \beta < 0$, there is the upper bound on $- \beta^{-1}$ as well.
Note that the previous work only considered the lower bound as $\beta$ was limited to be positive.
Under such boundaries, another parameter $\eta \in (-1, 1)$, which is invertible with $\beta$, is defined as the ratio between either of them and the value at the empirical worst-case TD error, $\pm \overline{|\delta|}$.
Specifically, $\beta_i$ is invertibly mapped from $\eta_i$ as follows:
\begin{align}
    \beta_i = - \cfrac{\mathrm{sgn}(\eta_i)}{\overline{|\delta|_i}} \ln(1 - |\eta_i|)
    \label{eq:eta2beta}
\end{align}
where $\mathrm{sgn}(\cdot)$ denots the sign function, and therefore, $\eta=0$ gets $\beta=0$.
As $\overline{|\delta|}$ is in the denominator, the scale issue is resolved.
In addition, as $\eta$ is in $(-1, 1)$, $\boldsymbol{\eta} = [\eta_1, \eta_2, \ldots, \eta_N]$ can approximately yield the regular intervals.

\subsection{Policy improvement with central value}

Since this algorithm estimates multiple value functions with the corresponding $\beta$, $\boldsymbol{V} = [V_1, V_2, \ldots, V_N]$, it can be interpreted as a kind of multi-objective RL \citep{hayes2022practical,ilboudo2023domains}.
That is, although these value functions originally represent the same objective (i.e. the expected return), in reality, an optimistic or pessimistic bias is added to each during learning.
Therefore, the objective represented by $i$-th value function can be regarded as $\mathbb{E}_{p_e, \pi}[\sum_k^\infty \gamma^k r_{t+k}^{\beta_i} \mid s_t = s]$, where $r^{\beta_i}$ the implicitly modified reward function, from the original one by $\beta_i$.
As a result, the policy aims to maximize the value functions representing multiple objectives, exactly in line with the multi-objective RL: if $\beta_i \simeq \beta_j$, $i$- and $j$-th objectives would have positive correlation; otherwise, e.g. $\beta_i \beta_j \ll 0$, they would have negative correlation.

Accordingly, two approaches are raised for acquiring the optimal policy: one is to optimize a single policy with the scalarized $\boldsymbol{V}$; and another is to build a mixture distribution, the components of which are optimized with the corresponding value functions.
The former makes the learning cost minimal while its policy depends on the way of scalarization and has no adaptability after learning.
The latter needs the $N$-times learning cost while its policy can be fine-tuned by adjusting its mixture ratio.
This paper does not focus on the way of fully exploiting the distributional value function, so for simplicity, the former is employed.

\begin{figure}[tb]
    \centering
    \includegraphics[keepaspectratio=true,width=0.96\linewidth]{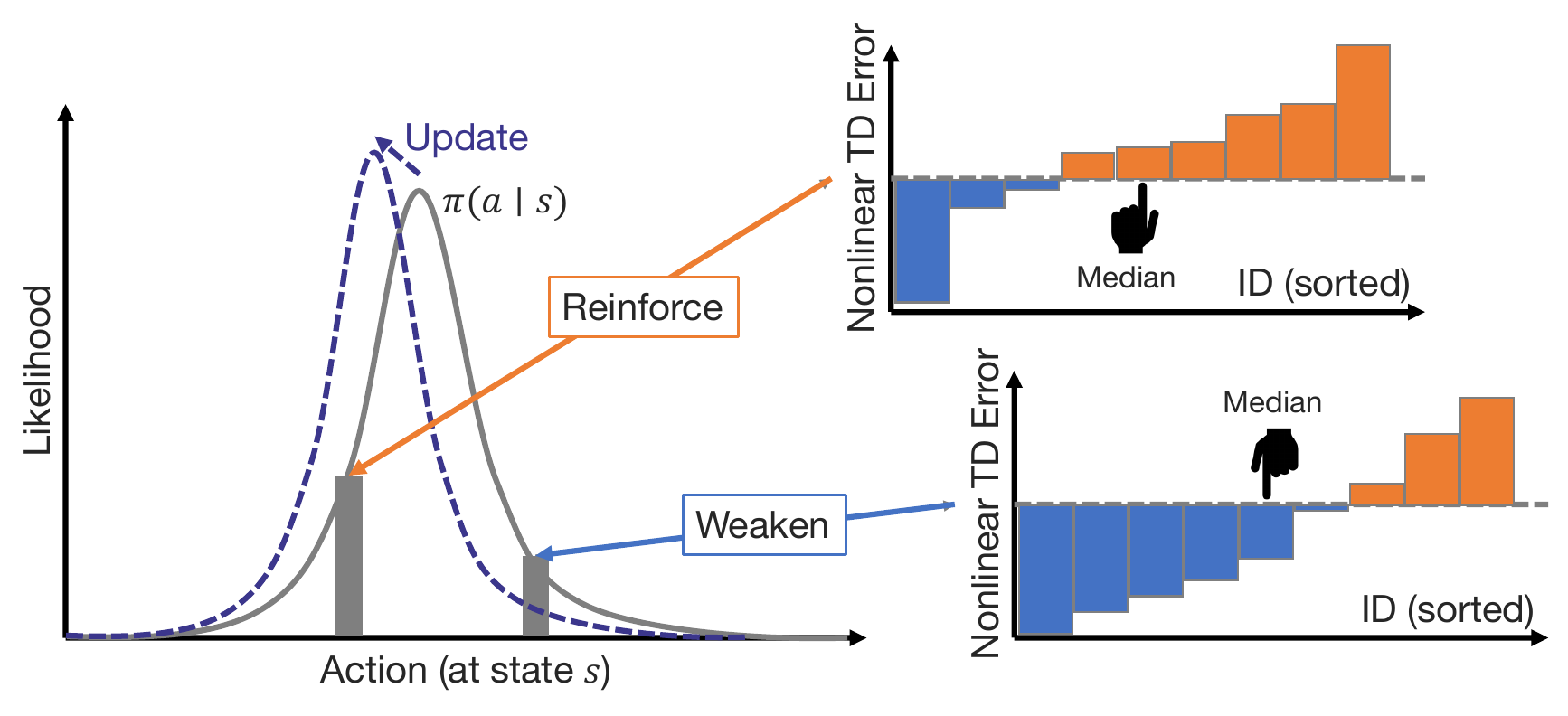}
    \caption{Policy improvement with ensemble of optimistic/pessimistic value functions:
    since learning multiple policies is costly, a single policy is optimized based on a central value of nonlinear TD errors calculated from multiple value functions;
    by employing the median as this central value, the direction of the policy improvement can be properly determined without being affected by outliers.
    }
    \label{fig:policy_scalar}
\end{figure}

Specifically, from $\boldsymbol{V}$ and $\boldsymbol{\beta}$, the respective nonlinear TD errors are computed as follows:
\begin{align}
    f_{\boldsymbol{\beta}}(\boldsymbol{\delta}) = [f_{\beta_1}(\delta_1), f_{\beta_2}(\delta_2), \ldots, f_{\beta_N}(\delta_N)]
\end{align}
The central value of $f_{\boldsymbol{\beta}}(\boldsymbol{\delta})$ is extracted using $\mathcal{M}: \mathbb{R}^N \to \mathbb{R}$, which replaces $f(\delta)$ in eq.~\eqref{eq:grad_pol}.
That is, the update rule of $\phi$ can be renewed as follows:
\begin{align}
    \begin{cases}
        \phi &\gets \phi - \alpha \mathbb{E}_{(s, a, s^\prime, r) \sim \mathcal{D}}[
        -\mathcal{M}(f_{\boldsymbol{\beta}}(\boldsymbol{\delta})) \nabla_\phi \ln \pi(a \mid s)]
        \\
        \mathcal{M}(f_{\boldsymbol{\beta}}(\boldsymbol{\delta})) &\overset{\mathrm{e.g.}}{=} \mathrm{Median}(f_{\boldsymbol{\beta}}(\boldsymbol{\delta}))
    \end{cases}
    \label{eq:policy_dist}
\end{align}
where the median operation is employed as the central value since it is robust to outliers in $f_{\boldsymbol{\beta}}(\boldsymbol{\delta})$ (see Fig.~\ref{fig:policy_scalar}).
As reducing excessive policy gradients makes RL stable \citep{schulman2017proximal}, this implementation, which ignores them as outliers, should enhance the generality of the proposed method.

In addition, the median operation has the potential to implicitly select an optimistic, pessimistic, or neutral policy according to the faced task and situation.
As can be seen in Fig.~\ref{fig:op_tderr}, $f_{\beta_1}(\delta) \leq f_{\beta_2}(\delta)$ ($\beta_1 < \beta_2$) holds, indicating that the neutral policy is expected only if $\boldsymbol{\delta}$ are approximately the same.
Conversely, since that condition is generally not guaranteed, optimistic or pessimistic bias is expected to occur.
In other words, the median operation is a simple, yet effective scalarization for utilizing the multiple value functions.

As a remark, in a prioritized experience replay (PER) \citep{schaul2016prioritized}, the absolute value of the (nonlinear) TD error is utilized for computing the priority of each experience data.
$f_{\boldsymbol{\beta}}(\boldsymbol{\delta})$ cannot be utilized for that purpose since the priority must be scalar if the buffer is shared for computing all loss functions, although the nonlinear TD error makes sampling optimistic or pessimistic.
Instead, the absolute value of $\mathcal{M}(f_{\boldsymbol{\beta}}(\boldsymbol{\delta}))$ is employed.

\section{Simulations}

\subsection{Conditions}

Here, experimental conditions about tasks, baselines, and learning implementations, are described.

\subsubsection{Task}

\begin{figure}[tb]
    \centering
    \includegraphics[keepaspectratio=true,width=0.96\linewidth]{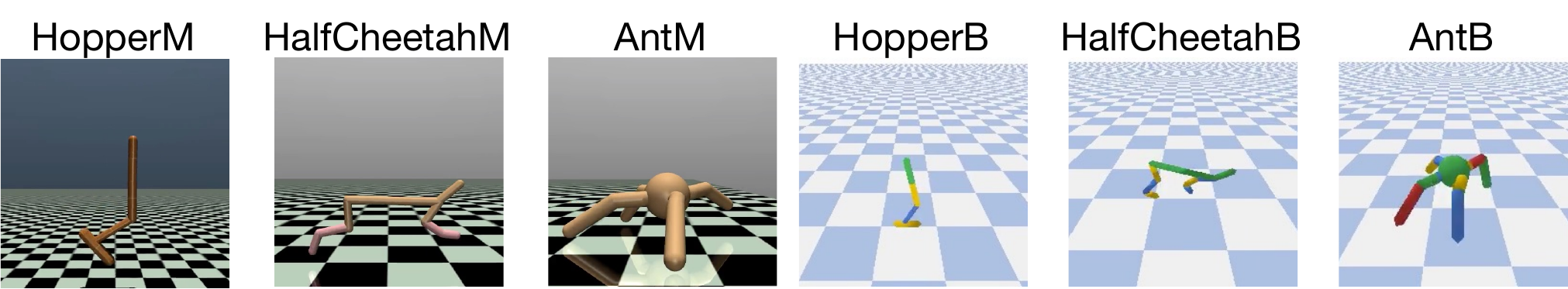}
    \caption{Benchmark tasks to be conducted in this paper:
    Mujoco \citep{todorov2012mujoco} and Pybullet \citep{coumans2016pybullet} for simulating three tasks are distinguished with `M' and `B' on the suffix of each task, respectively;
    it should be noticed that in these two simulators, the dynamics, models and reward functions are different.
    }
    \label{fig:task}
\end{figure}

The proposed method, DROP, is verified on the learning performance of a total of six tasks.
Specifically, Hopper, HalfCheetah, and Ant implemented respectively on Mujoco \citep{todorov2012mujoco} and Pybullet \citep{coumans2016pybullet} are employed as the tasks, as shown in Fig.~\ref{fig:task}.
Here, `M' for Mujoco and `B' for Pybullet are added to the suffix of each task name to distinguish between them.
The reason why two simulators are tested is that their dynamics computation, models used (especially their maximum torques), and reward functions are different, resulting in different learning tendencies.
Note that there are several versions of task implementation on Mujoco, but v4 is used in this paper.
In addition, the OpenAI Gym API is used, and unlike the new API in Gymnasium, when a task is terminated due to a time limit, it is treated in the same way as an end condition defined for the task.

All tasks are terminated in the number of episodes rather than in the number of data experienced, with 2000 episodes being the upper limit in this paper.
The learning performance (or score) is evaluated by the statistic (i.e. the interquartile mean referring to the literature \citep{agarwal2021deep}) of the returns obtained by 100 episodes with the optimized policy.
The 12 models for each method are initialized with different seeds and trained in the same way.
Then, the proposed and baseline methods are compared statistically.

\subsubsection{Heuristic model with asymmetric learning rates}

In the heuristic model proposed in the previous studies \citep{dabney2020distributional,muller2024distributional}, the learning rate $\alpha$ itself is switched by the positive or negative TD error, as $\alpha^{+,-}$, as illustrated in Fig.~\ref{fig:background}.
This implementation corresponds to the expectile regression of value distribution, and  differs from DROP, making a fair comparison difficult.
This paper, therefore, modifies it slightly to include a learning rate multiplier (i.e. $\alpha^{+,-}/\alpha$) in $f(\delta)$, in line with this study.
\begin{align}
    f^{\mathrm{heuristic}}_\eta(\delta) = (1 + \mathrm{sgn}(\delta) \eta) \delta
    \label{eq:heuristic}
\end{align}
where $\eta \in (-1, 1)$ is the optimistic/pessimistic hyperparameter as well as that of DROP.
That is, with $\eta \to 1$, $\delta < 0$ is ignored for optimistically updating $\theta$ (and $\phi$); and with $\eta \to -1$, the perfect pessimism is obtained by ignoring $\delta > 0$.

\subsubsection{Learning}

To stabilize RL and make the experience replay available, ERC \citep{kobayashi2024revisiting} is employed in conjunction with DROP.
Therefore, the learning conditions including the network architecture are basically based on the implementations in that literature.
For example, by using AdaTerm \citep{ilboudo2023adaterm} with its default setting (e.g. $\alpha=10^{-3}$) as an optimizer, the network parameters, $\theta$ and $\phi$, are learned robustly to the noise of (nonlinear) TD errors.
In addition, each value function with $\beta_i$ (or $\eta_i$) ($i=1,\ldots,N$) is accurately estimated with the ensemble model \citep{kobayashi2023reward}, which has $K$ heads with trainable and randomly-fixed weights for computing statistic of their outputs.

However, the use of the replay buffer is changed from the usual uniform sampling to PER \citep{schaul2016prioritized} because the use of PER can emphasize optimism (and pessimism in this study), as indicated in the previous work \citep{kobayashi2022optimistic}.
In that case, the batch size is set to 32 and the number of experiences to be replayed at the end of the episode is set to $1/8$ of the buffered data (a maximum capacity of 102400).
In addition, as for the number of ensembles, $N$, and the range of optimism and pessimism, $\boldsymbol{\eta}$, in the critic, they are additional elements that appeared in this study and are tuned as shown in the next section.

The overall learning conditions are summarized in Table~\ref{tab:param}.
Note that hyperparameters used in the combined methods are left at their default settings unless otherwise explained.

\begin{table*}[tb]
    \caption{Learning conditions:
    the methods combined in the proposed algorithm use the default hyperparameters given in the references presented, unless otherwise noted.
    }
    \label{tab:param}
    \centering
    {\scriptsize
    \begin{tabular}{lc}
        \hline\hline
        \#Hidden layers of fully connected networks & $2$
        \\
        \#Neurons for each hidden layer & $100$
        \\
        Activation function & Squish \citep{barron2021squareplus,kobayashi2023design}
        \\
        & + RMSNorm \citep{zhang2019root}
        \\
        Estimation model of each value function & Ensemble model \citep{kobayashi2023reward}
        \\
        Distribution model of policy & Student's t-distribution \citep{kobayashi2019student}
        \\
        \hline
        Discount factor $\gamma$ & $0.99$
        \\
        Optimizer & AdaTerm \citep{ilboudo2023adaterm}
        \\
        The way of updating target networks & CAT-soft update \citep{kobayashi2024consolidated}
        \\
        Stabilization tricks & L2C2 \citep{kobayashi2022l2c2} + ERC \citep{kobayashi2024revisiting}
        \\
        \hline
        Buffer size $|\mathcal{D}|$ & $102400$
        \\
        Batch size & $32$
        \\
        \#Replayed data & $|\mathcal{D}|/8$
        \\
        Hyperparameters of PER \citep{schaul2016prioritized} $(\alpha^{\mathrm{PER}}, \beta^{\mathrm{PER}})$ & $(1, 0.5)$
        \\
        \hline\hline
    \end{tabular}
    }
\end{table*}

\subsection{Results}

According to the above conditions, three experiments are conducted.

\subsubsection{Parameter tuning}

\begin{figure}[tb]
    \begin{subfigure}[b]{0.48\linewidth}
        \centering
        \includegraphics[keepaspectratio=true,width=\linewidth]{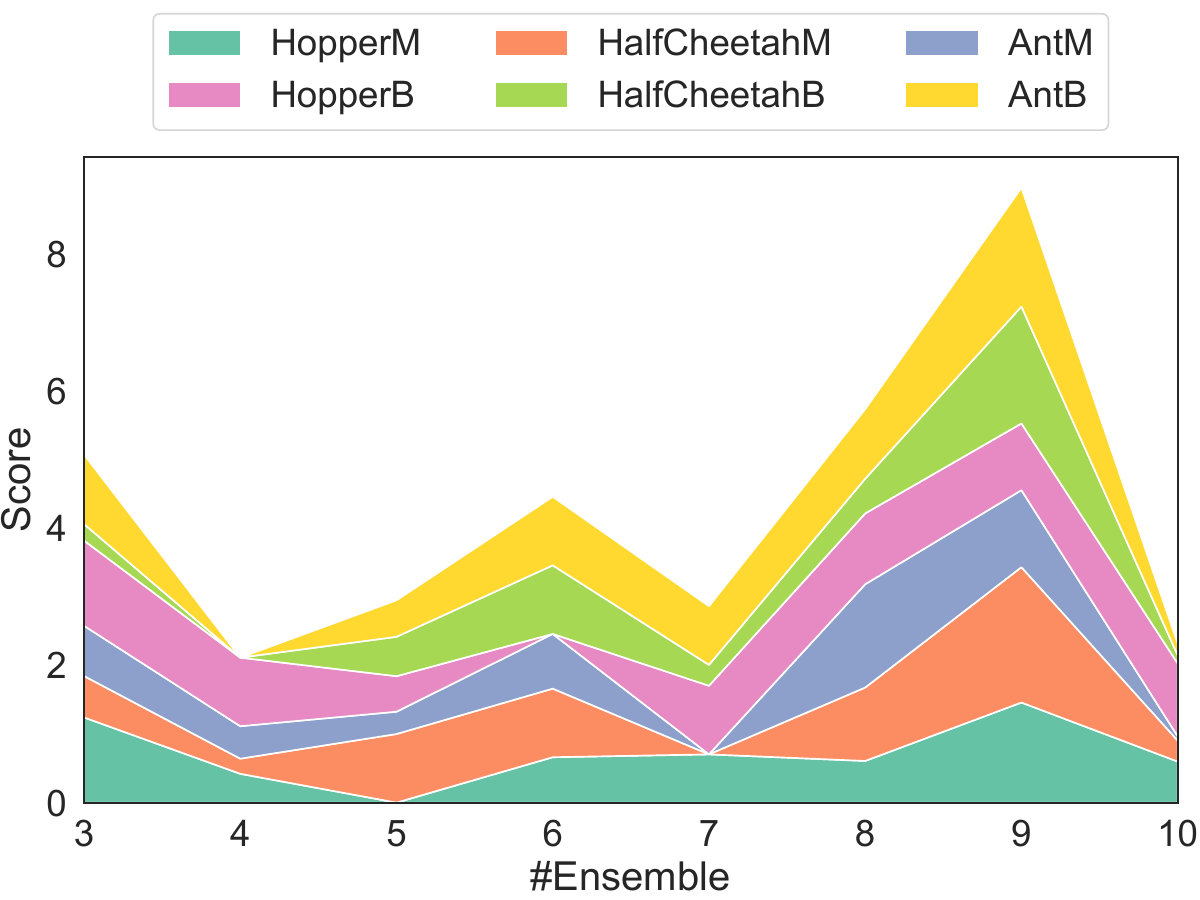}
        \subcaption{The number of ensemble models}
        \label{fig:result_ensemble}
    \end{subfigure}
    \begin{subfigure}[b]{0.48\linewidth}
        \centering
        \includegraphics[keepaspectratio=true,width=\linewidth]{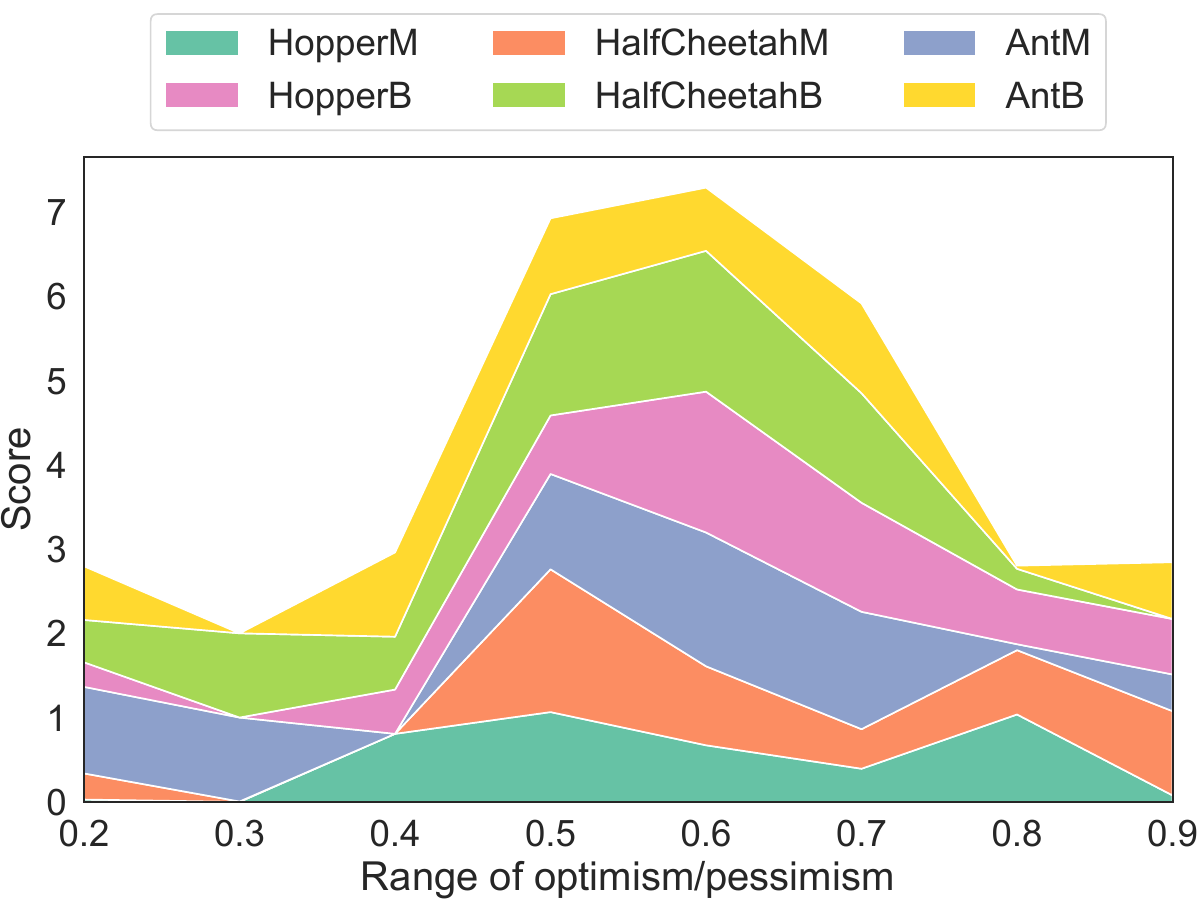}
        \subcaption{The range of optimism/pessimism}
        \label{fig:result_range}
    \end{subfigure}
    \caption{Parameter tuning in DROP:
    the score for each score is computed as the interquartile mean of 12 random seeds and normalized;
    the scores for each condition are divided by the gap between their maximum and minimum in order to consider generality to tasks;
    as a result, $N=9$ and $\eta_i \in [-0.6, 0.6]$ obtained the best score.
    }
    \label{fig:result_parameter}
\end{figure}

First, the number of ensembles, $N$, is tuned by a gridsearch.
Here, the maximum of $\boldsymbol{\eta}$ is tentatively set to a moderate $0.5$ as suggested by the previous study \citep{kobayashi2022optimistic}, and the minimum is set to $-0.5$ as well.
Then, $\boldsymbol{\eta}$ is regularly placed according to $N=3,4,\ldots,10$ using a linspace function.

The score of each task is normalized to $[0, 1]$, and in order to consider generality to tasks, it is divided by the gap between the maximum and minimum scores in each condition.
As shown in Fig.~\ref{fig:result_ensemble}, the best score was obtained for $N=9$.
Although the trend seems to be that the score improves with larger $N$, the score with $N=10$ was suddenly deteriorated.
This may be due to the fact that the ensemble model was not implemented in independent networks but was partially shared, as shown in Fig.~\ref{fig:value_dist}, resulting in a lack of expressiveness.
This is also consistent with the fact that the slight improvement in performance at $N=3$, probably due to the extra expressiveness.

Next, the maximum and minimum values of $\boldsymbol{\eta}$, $\pm \bar{\eta}$, are tuned by a gridsearch along with the tuned $N=9$.
$\bar{\eta} = 0.2, 0.3, \ldots, 0.9$ with $0.1$ increments are compared as in the case of $N$.
The scores are shown in Fig.~\ref{fig:result_range}, where $\bar{\eta}=0.6$ is the best one.
This result is consistent with the previous study, which showed that a moderate level of optimism (and pessimism) is superior to the others.

\subsubsection{Comparison}
\label{subsubsec:comparison}

\begin{figure}[tb]
    \centering
    \includegraphics[keepaspectratio=true,width=0.96\linewidth]{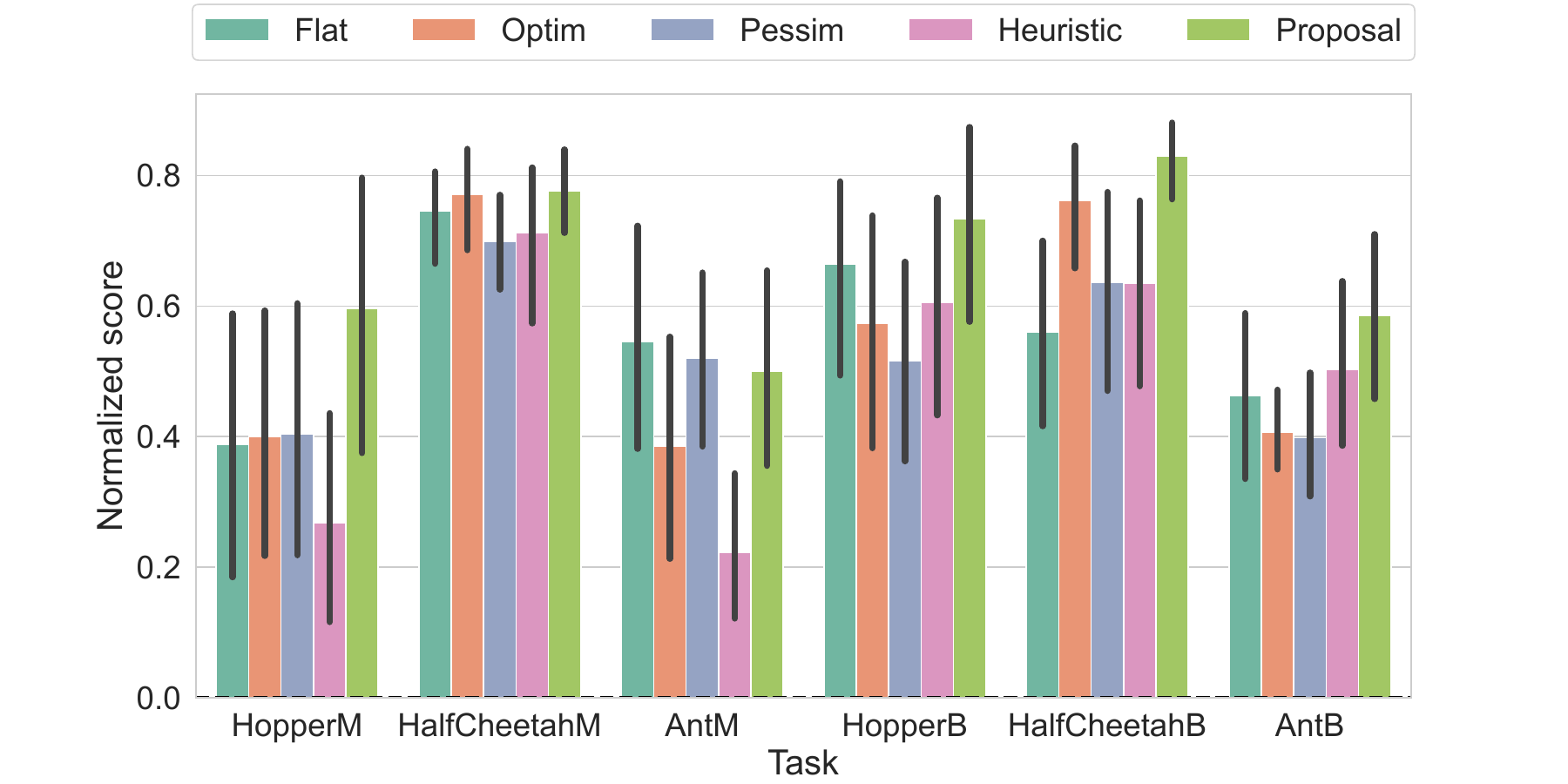}
    \caption{Comparision of five methods:
    the first three methods, \textit{Flat}, \textit{Optim}, and \textit{Pessim}, represent the cases with $\eta=0, 0.6, -0.6$ and $N=1$, respectively, \textit{Heuristic} is based on the previous studies \citep{dabney2020distributional,muller2024distributional} implemented in eq.~\eqref{eq:heuristic}, and \textit{Proposal} means the proposed method, named DROP;
    only \textit{Proposal} succeeded in all the tasks stably.
    }
    \label{fig:result_comp}
\end{figure}

From the parameter tuning conducted above, $N=9$ with the range of $\pm0.6$, i.e. $\boldsymbol{\eta} = [-0.6, -0.45, -0.3, -0.15, 0, 0.15, 0.3, 0.45, 0.6]$, is selected for the proposed DROP.
As well, the heuristic model defined in eq.~\eqref{eq:heuristic} employs the same setting for comparison.
With three additional ablation tests, the following five methods are compared in total.
\begin{itemize}
    \item \textit{Flat}: with $\boldsymbol{\eta} = [0]$ as a non-biased traditional RL (equal to the baseline, ERC)
    \item \textit{Optim}: with $\boldsymbol{\eta} = [0.6]$ as an optimistic RL derived in the previous study \citep{kobayashi2022optimistic}
    \item \textit{Pessim}: with $\boldsymbol{\eta} = [-0.6]$ as a pessmistic RL derived in this paper
    \item \textit{Heuristic}: with the tuned parameters and the replacement of eq.~\eqref{eq:drop} with eq.~\eqref{eq:heuristic} as a heuristic model in the literature \citep{dabney2020distributional,muller2024distributional}
    \item \textit{Proposal}: with the tuned parameters as the proposed DROP
\end{itemize}

The scores are summarized in Fig.~\ref{fig:result_comp}.
Note that the corresponding 95~\% confidence intervals are provided for reference, since the scores vary greatly depending on the success or failure of the task and are bimodal.
Further analysis of the results is presented in Appendix~\ref{app:performance}.

First, \textit{Flat}, \textit{Optim}, and \textit{Pessim} excelled at different tasks.
For example, \textit{Optim} with better exploration capability was superior for HalfCheetahB, which tends to stay in a local solution of standing due to insufficient torques, while \textit{Pessim} with better safety capability and \textit{Flat} were superior for AntM, which is prone to fall over and gain unworthy experience due to excessive torques.
In the latter case, the reason why \textit{Flat} was also superior is probably due to the fact that the recent RL stabilization tricks (e.g. the ones used in this paper as summarized in Table~\ref{tab:param}) is to make learning proceed conservatively, i.e. pessimistically.
This reason is also reasonable for explaining the reason why \textit{Pessim} was not good in the other tasks, namely, the overlapping effects prevented the sufficient exploration for finding a global optimum.

\begin{figure}[tb]
    \begin{subfigure}[b]{0.48\linewidth}
        \centering
        \includegraphics[keepaspectratio=true,width=\linewidth]{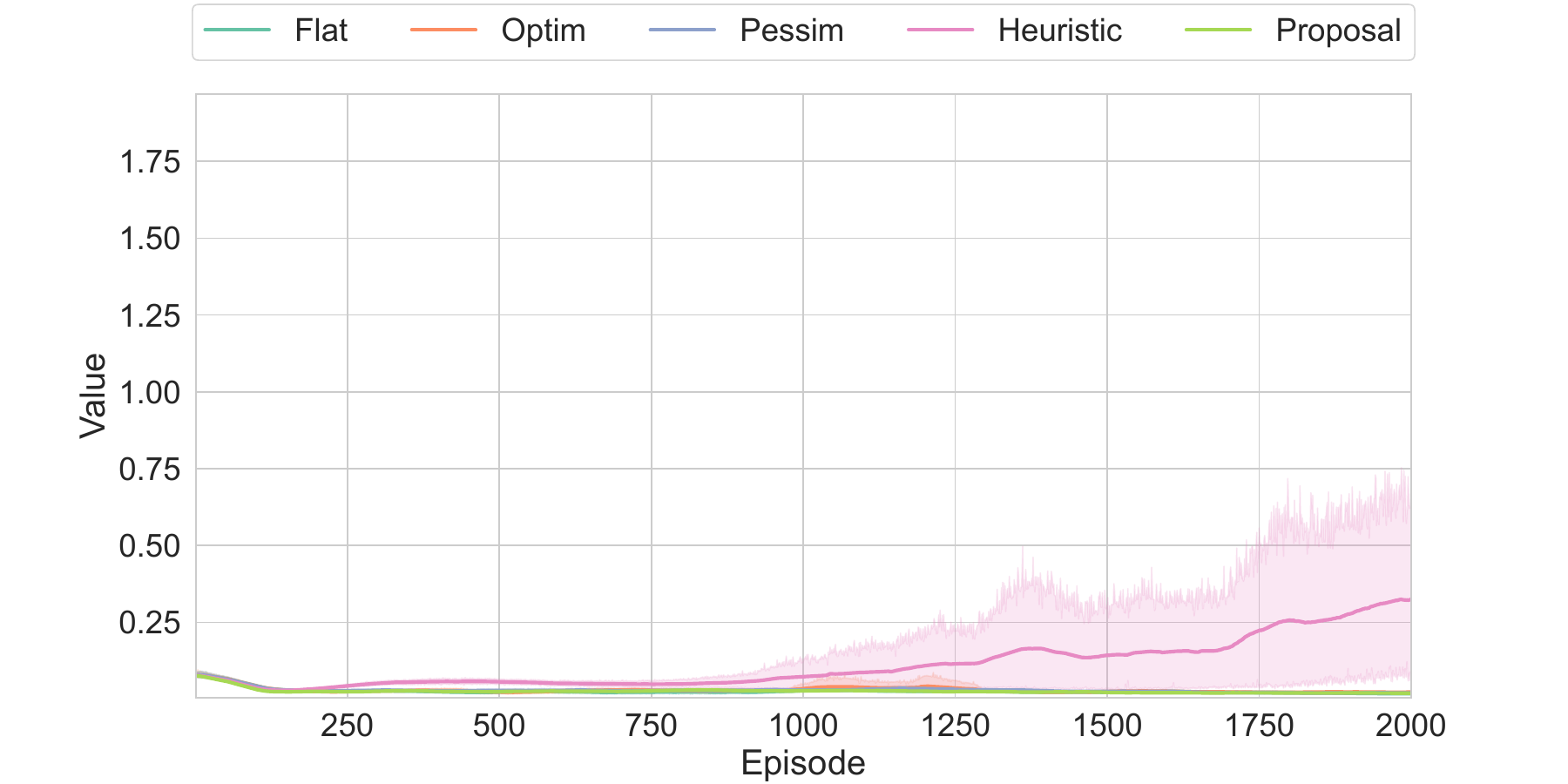}
        \subcaption{HopperM}
        \label{fig:tderr_scale_HopperM}
    \end{subfigure}
    \begin{subfigure}[b]{0.48\linewidth}
        \centering
        \includegraphics[keepaspectratio=true,width=\linewidth]{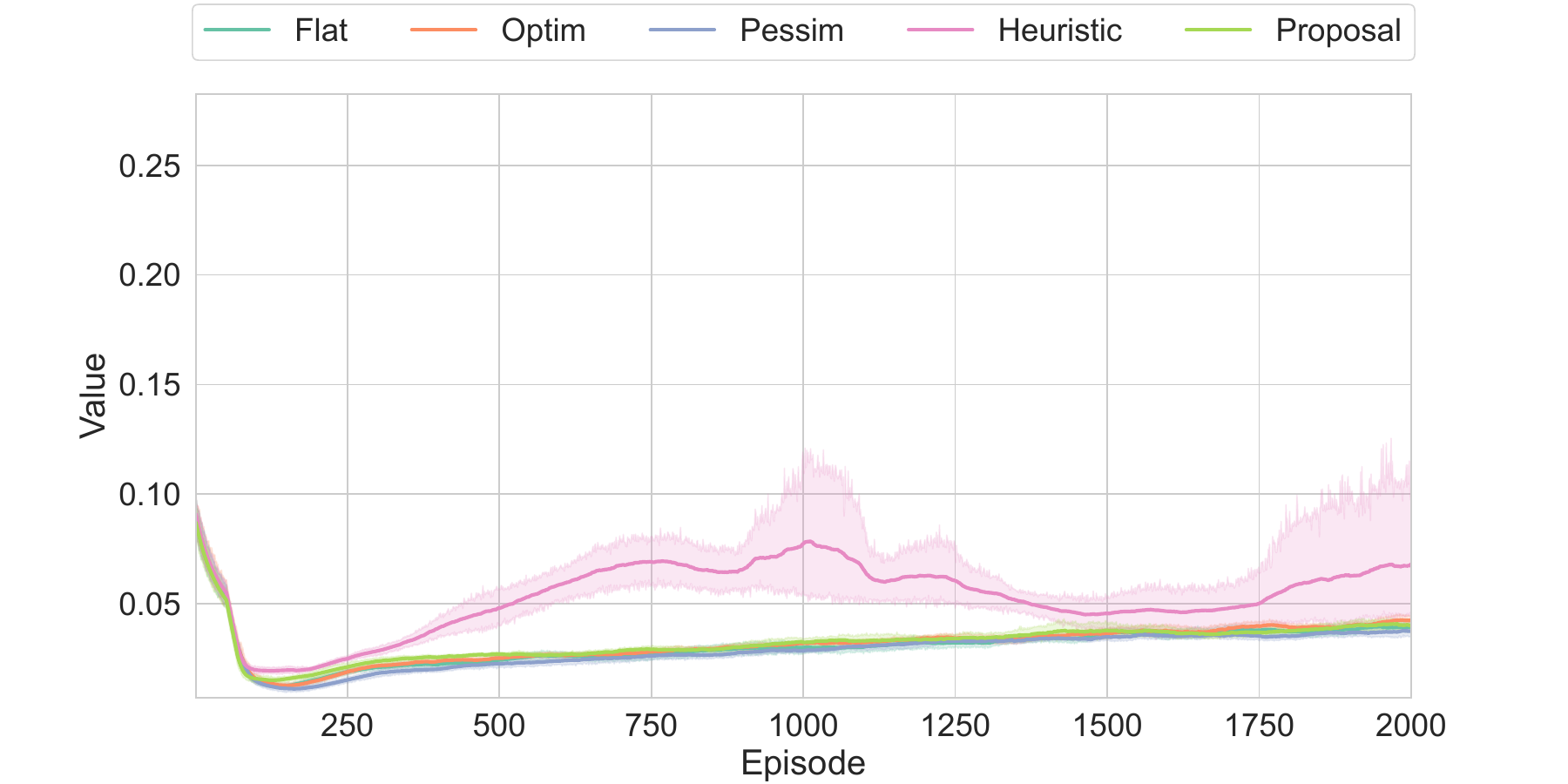}
        \subcaption{AntM}
        \label{fig:tderr_scale_AntM}
    \end{subfigure}
    \caption{The averaged scale of TD errors during learning:
    only \textit{Heuristic} failed the convergence of TD error in these tasks, collapsing the policy as can be seen in Fig.~\ref{fig:result_comp}.
    }
    \label{fig:tderr_scale}
\end{figure}

\begin{figure}[tb]
    \begin{subfigure}[b]{0.48\linewidth}
        \centering
        \includegraphics[keepaspectratio=true,width=\linewidth]{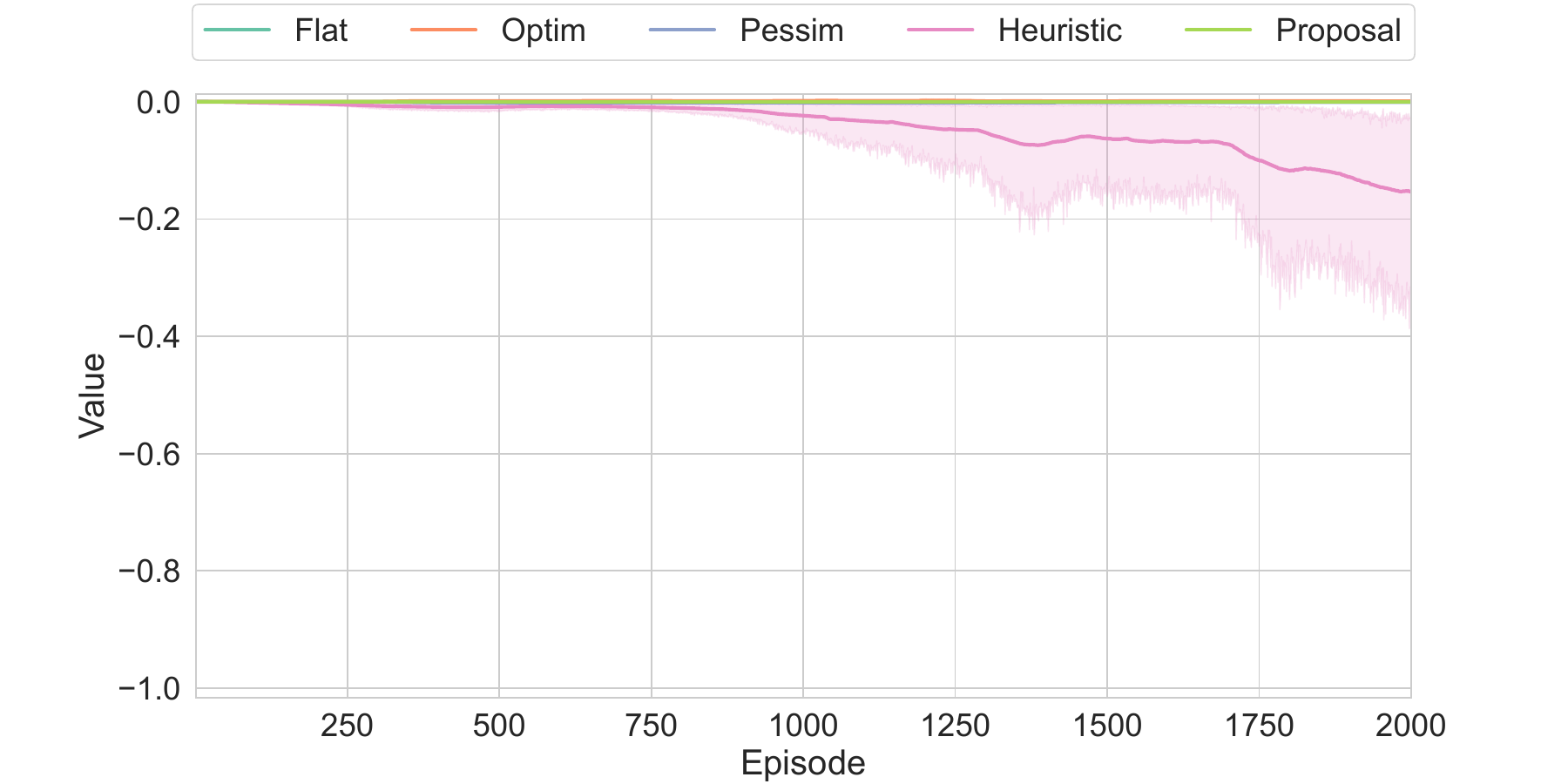}
        \subcaption{HopperM}
        \label{fig:tderr_diff_HopperM}
    \end{subfigure}
    \begin{subfigure}[b]{0.48\linewidth}
        \centering
        \includegraphics[keepaspectratio=true,width=\linewidth]{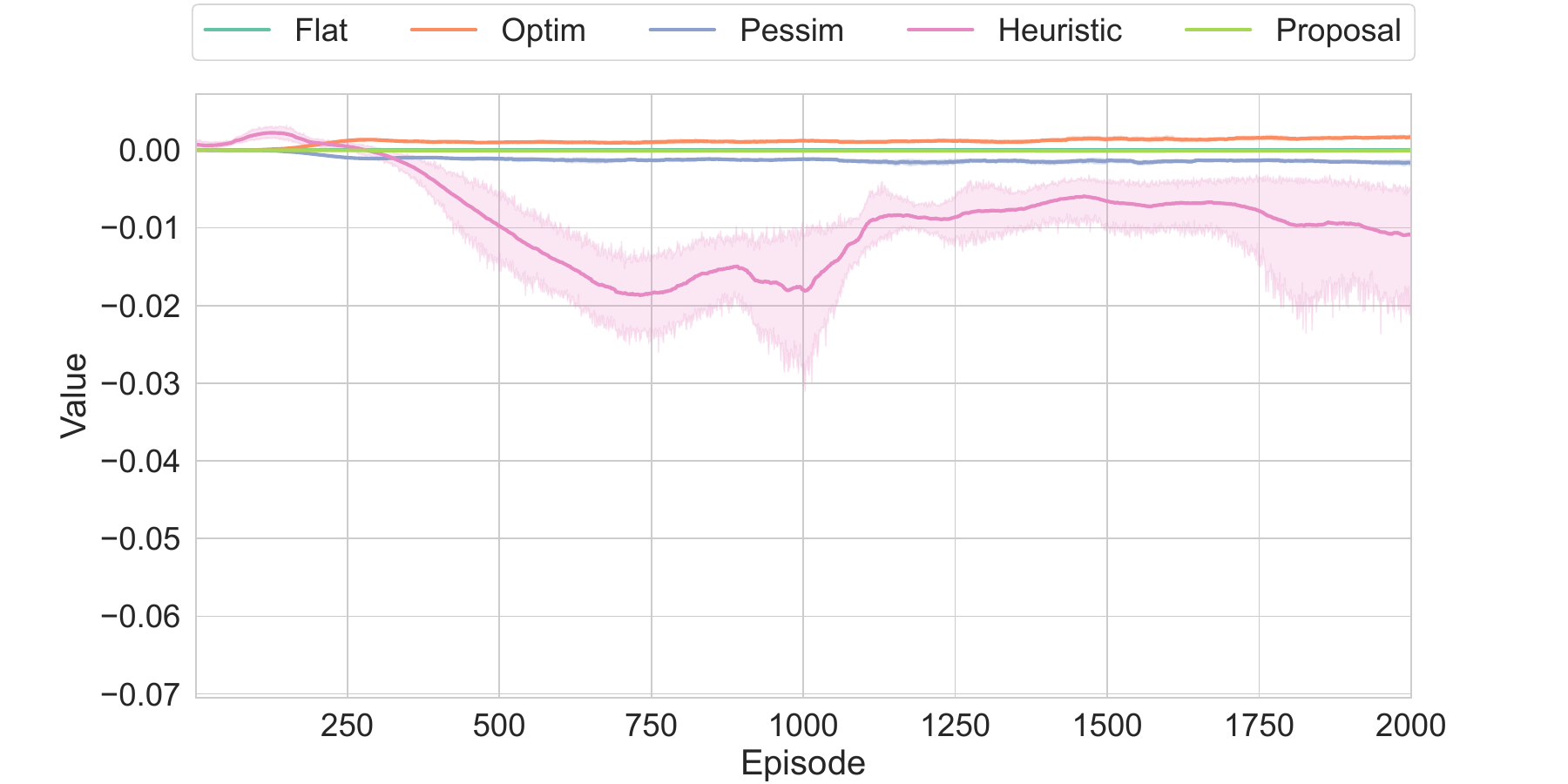}
        \subcaption{AntM}
        \label{fig:tderr_diff_AntM}
    \end{subfigure}
    \caption{The averaged optimistic/pessimistic bias during learning:
    in \textit{Heuristic}, the TD error for the pessimistic value function was not minimized appropriately since the bias diverged negatively;
    such a bias was not observed in \textit{Proposal} by canceling out the small biases in the optimistic and pessimistic value functions.
    }
    \label{fig:tderr_diff}
\end{figure}

\begin{figure*}[tb]
    \begin{subfigure}[b]{0.32\linewidth}
        \centering
        \includegraphics[keepaspectratio=true,width=\linewidth]{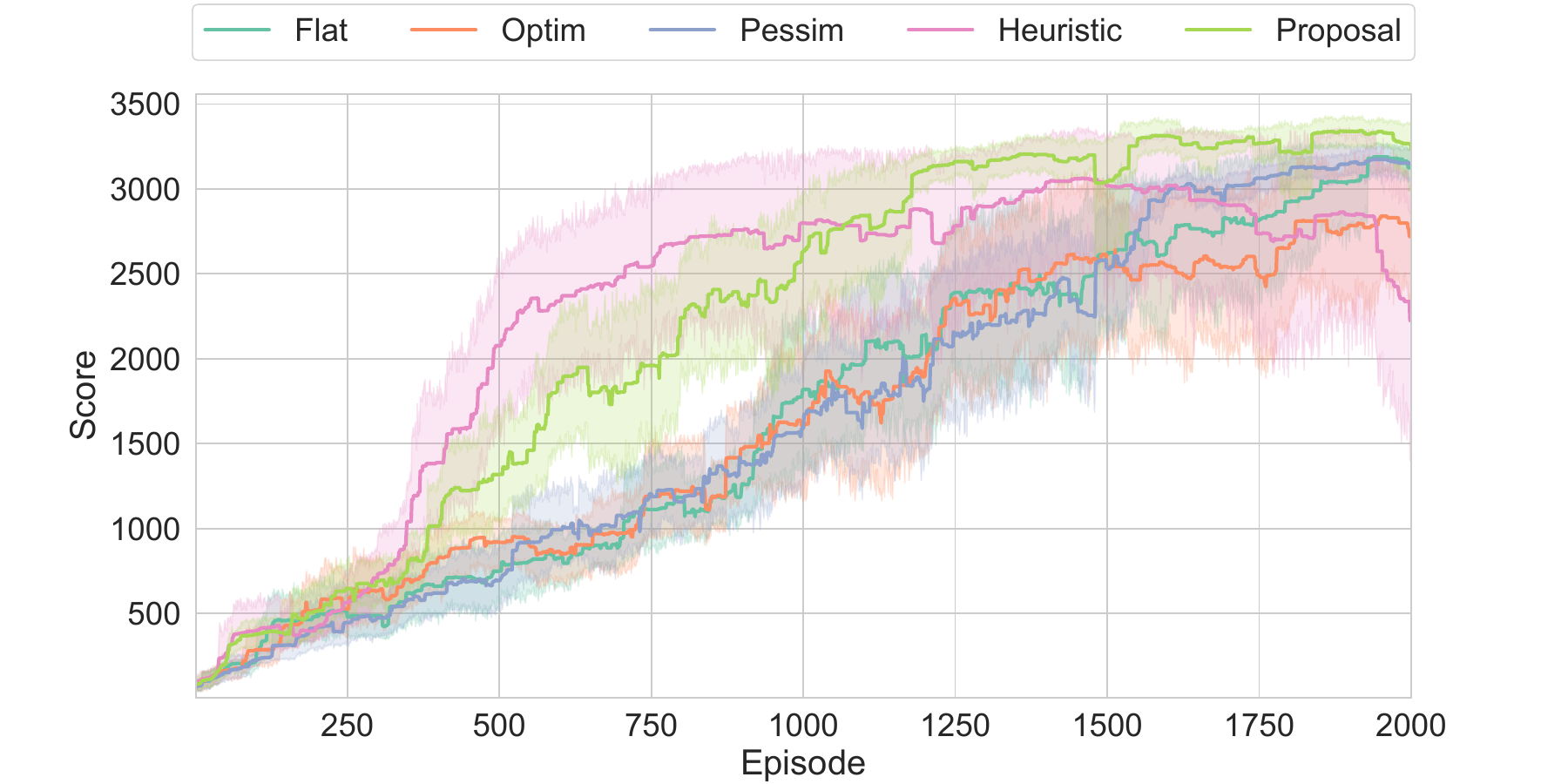}
        \subcaption{HopperM}
        \label{fig:score_HopperM}
    \end{subfigure}
    \begin{subfigure}[b]{0.32\linewidth}
        \centering
        \includegraphics[keepaspectratio=true,width=\linewidth]{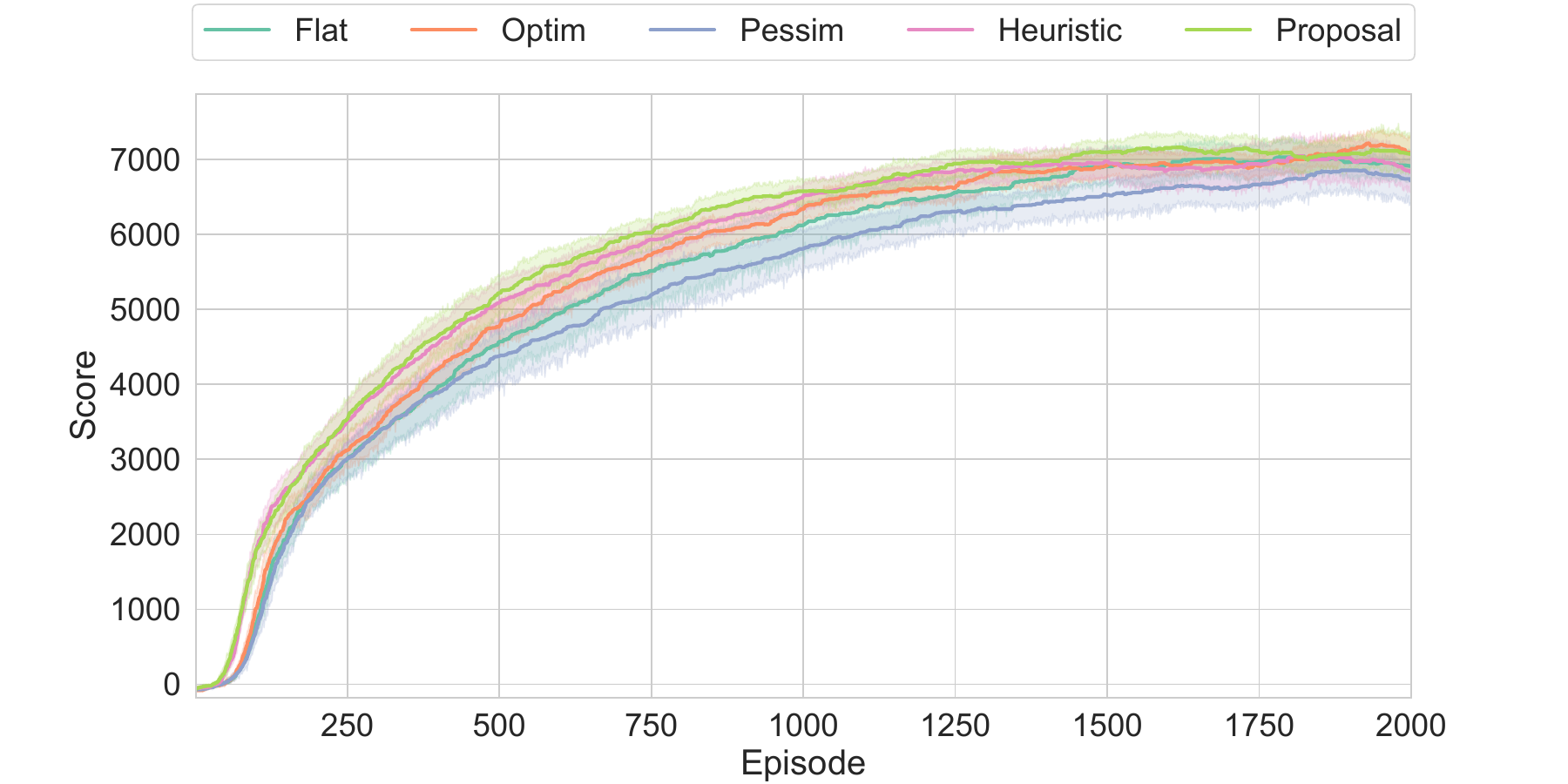}
        \subcaption{HalfCheetahM}
        \label{fig:score_HalfCheetahM}
    \end{subfigure}
    \begin{subfigure}[b]{0.32\linewidth}
        \centering
        \includegraphics[keepaspectratio=true,width=\linewidth]{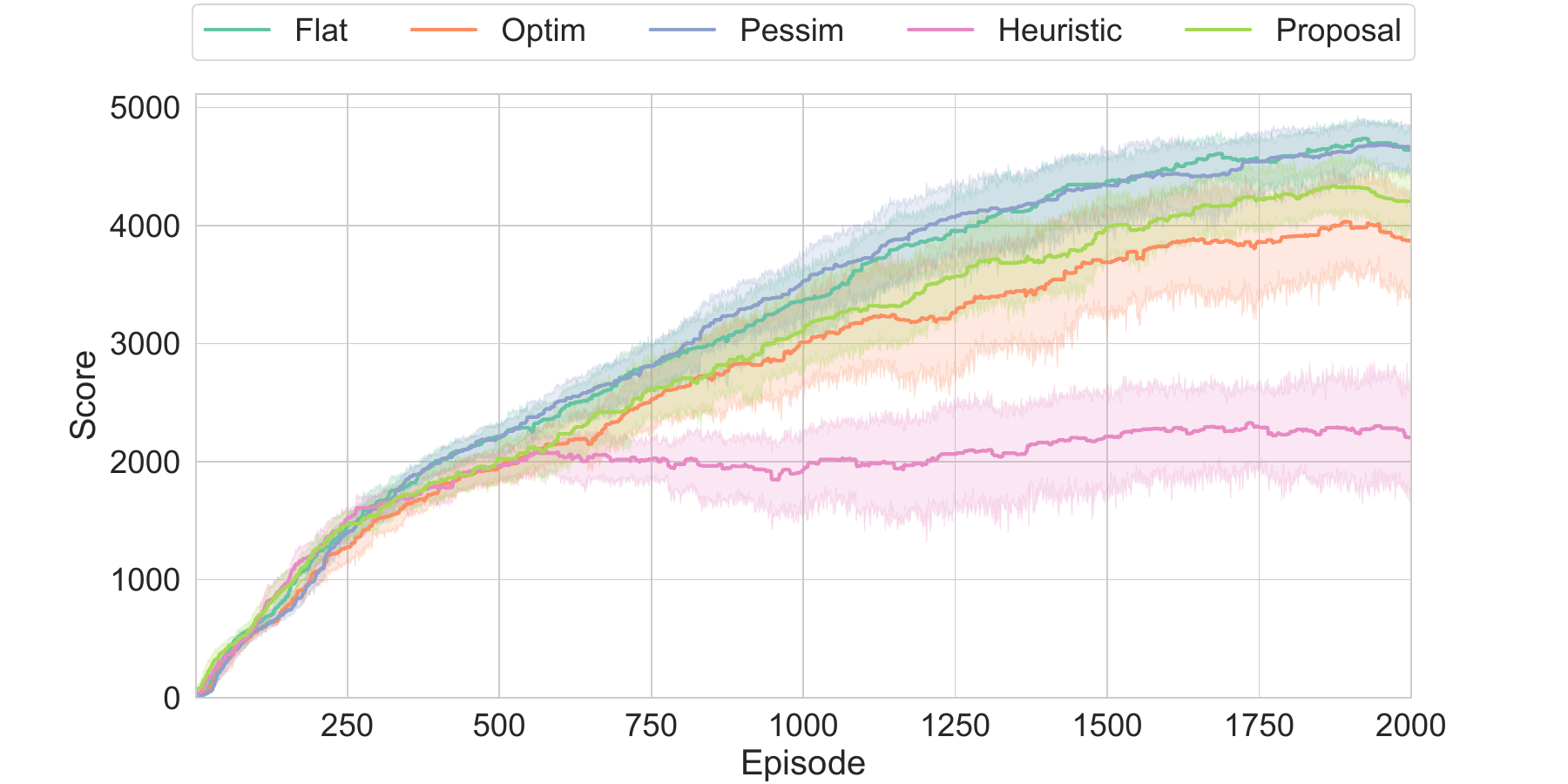}
        \subcaption{AntM}
        \label{fig:score_AntM}
    \end{subfigure}
    \begin{subfigure}[b]{0.32\linewidth}
        \centering
        \includegraphics[keepaspectratio=true,width=\linewidth]{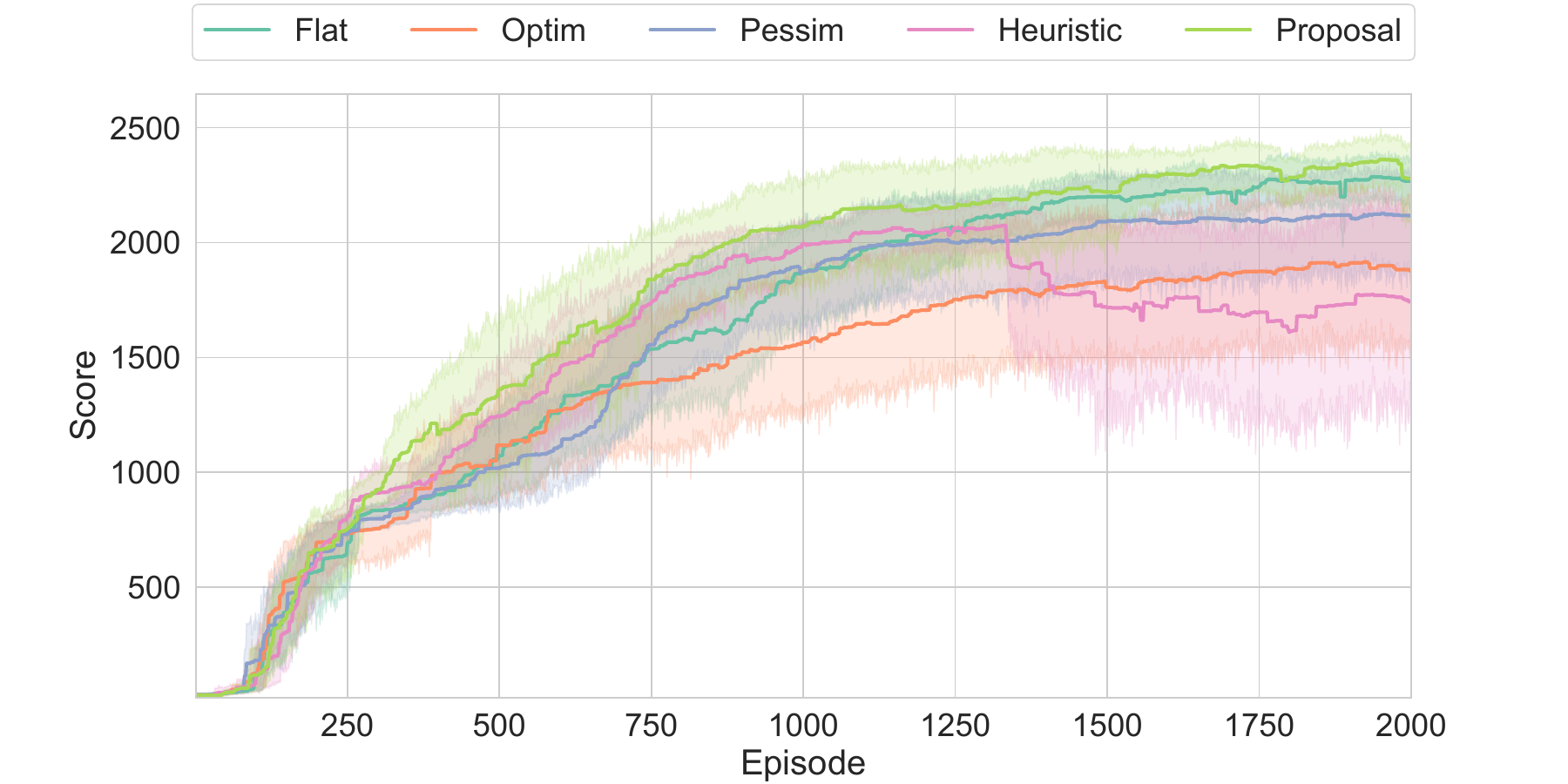}
        \subcaption{HopperB}
        \label{fig:score_HopperB}
    \end{subfigure}
    \begin{subfigure}[b]{0.32\linewidth}
        \centering
        \includegraphics[keepaspectratio=true,width=\linewidth]{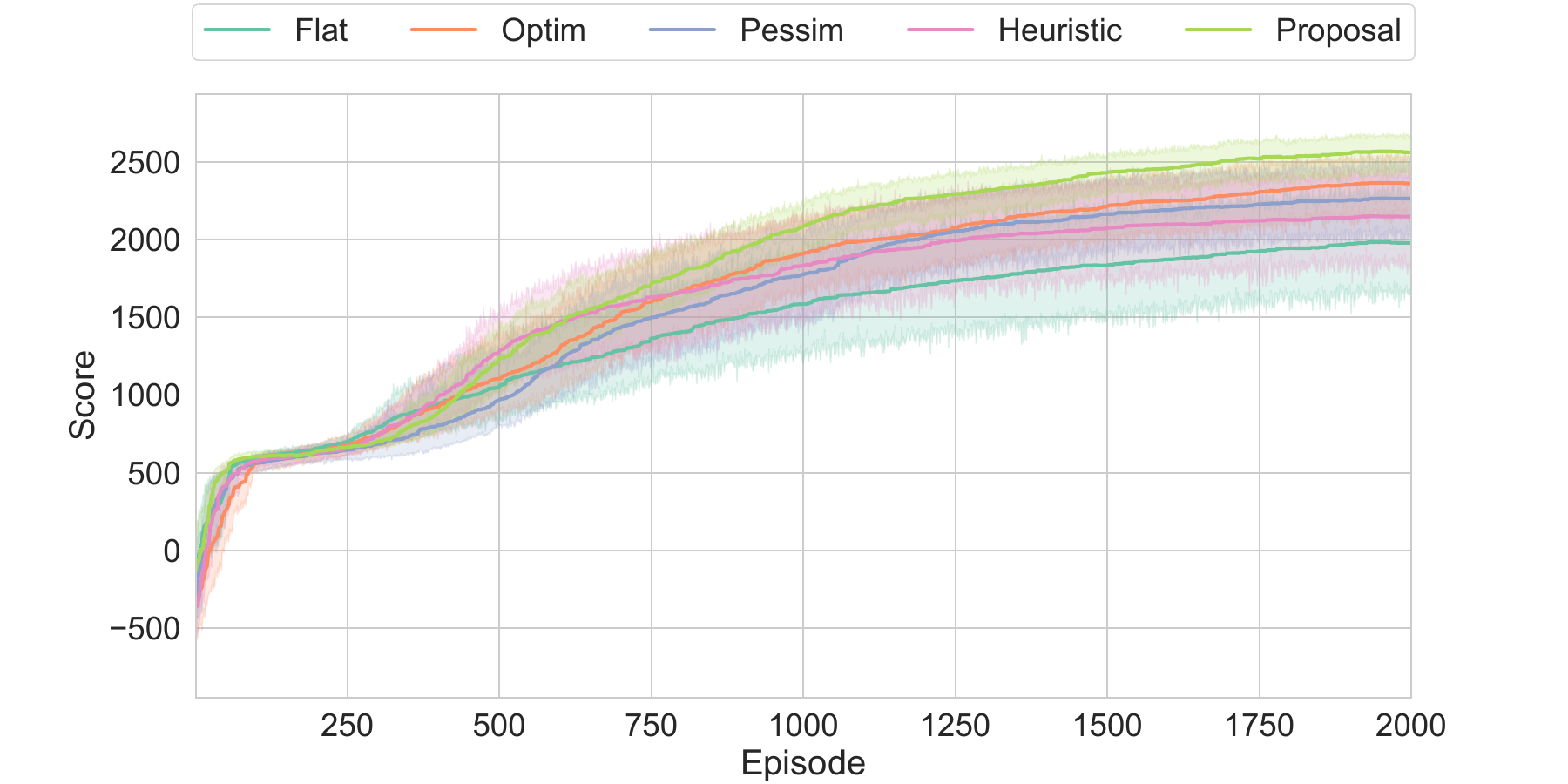}
        \subcaption{HalfCheetahB}
        \label{fig:score_HalfCheetahB}
    \end{subfigure}
    \begin{subfigure}[b]{0.32\linewidth}
        \centering
        \includegraphics[keepaspectratio=true,width=\linewidth]{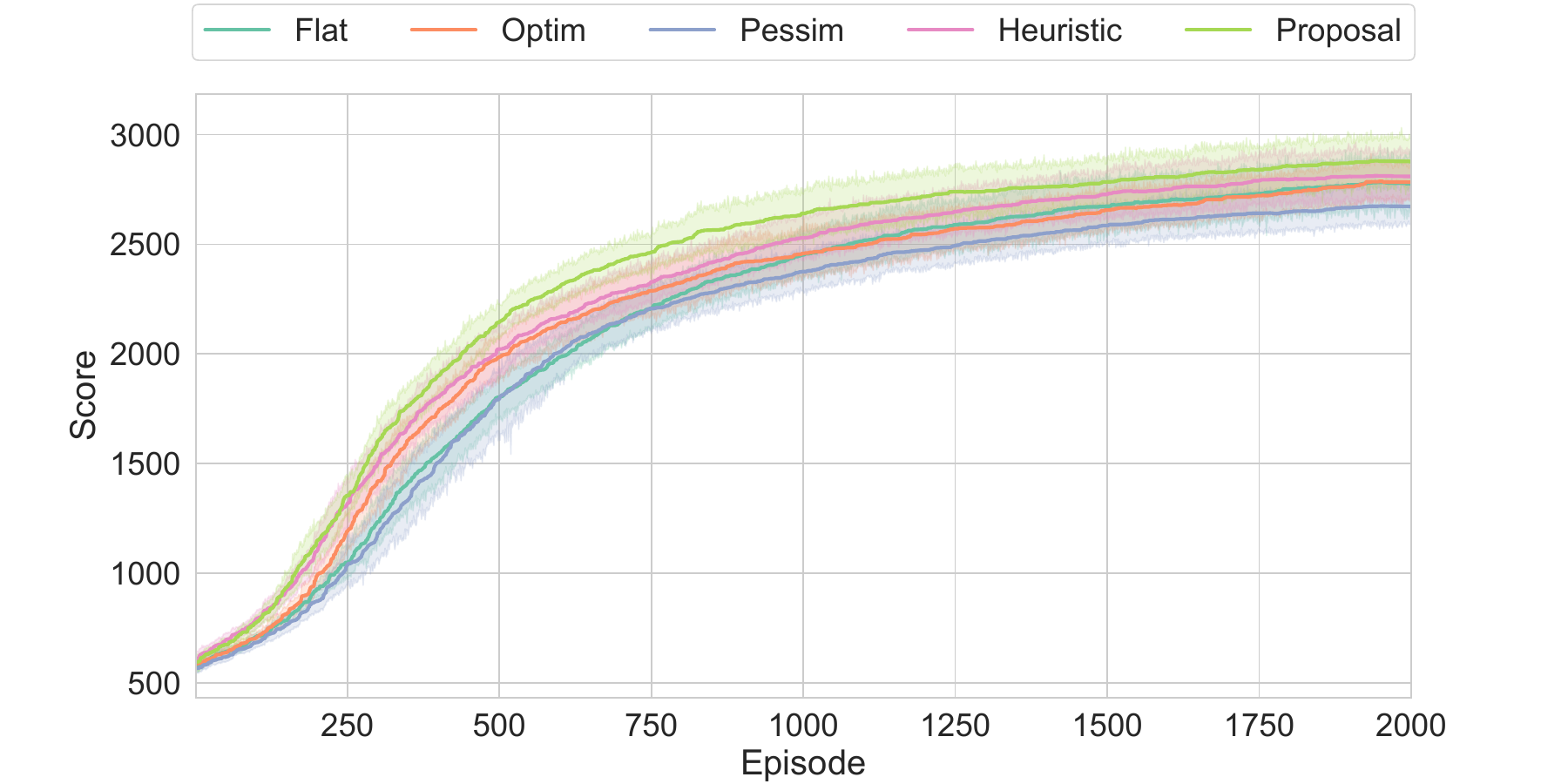}
        \subcaption{AntB}
        \label{fig:score_AntB}
    \end{subfigure}
    \caption{Learning curves of return:
    \textit{Heuristic} was unstable even if it increased the return rapidly, collapsing the policy suddenly;
    \textit{Proposal} was faster and more stable than others in most tasks.
    }
    \label{fig:score}
\end{figure*}

Next, the scores for \textit{Heuristic} were generally poor.
In particular, it have the worst scores on HopperM and AntM.
This should be due to the discontinuous TD error near $\delta \simeq 0$, destabilizing its learning.
In fact, looking at the learning curve w.r.t. the TD error scale in Fig.~\ref{fig:tderr_scale}, only \textit{Heuristic} was not able to make the TD error converge to almost zero.
In addition, as suggested in Fig.~\ref{fig:tderr_diff} for the averaged optimistic/pessimistic bias computed as $f(\delta) - \delta$, the pessimistic learning of the value function was unsuccessful, as it was biased negatively.
Note that \textit{Proposal} kept the bias zero, implying both the optimistic and pessimistic value functions were trained to the same level.

On the other hand, only \textit{Proposal} achieved high-quality behaviors in all tasks, namely, it outperformed or was comparable to scores of the other methods.
In fact, when the behaviors at the average scores were checked, as shown in YouTube\footnote{\url{https://youtu.be/5YdBbVQt7Q4}}, only its behaviors could be considered qualitatively successful in all tasks.
In addition, the learnings curve w.r.t. the returns are depicted in Fig.~\ref{fig:score}, indicating that it learned the optimal policies earlier than the other methods.

While the hyperparameters tuned with DROP in the above section may be one possible factor of this superiority in \textit{Proposal}, it cannot be the reason for collapsing the learning of \textit{Heuristic}, which is the main baseline for comparison.
The design by eq.~\eqref{eq:heuristic} of the heuristic model is also not much different in terms of the ratio of positive and negative updates: for the $\eta$ used in this paper, $4:1$ at most in the heuristic model, while DROP defined by eqs.~\eqref{eq:drop} and~\eqref{eq:eta2beta} has the similar ratio, $5:2$, when $\delta = \pm \overline{|\delta|}$.
Note that $\overline{|\delta|}$ is the empirically estimated maximum scale, so the actual TD error can cause $\delta > \overline{|\delta|}$ (or $\delta < -\overline{|\delta|}$), making the ratio larger.
Thus, it can be concluded that the theoretically-grounded DROP performed better than the heuristic model.

\subsection{Comparisions to the state-of-the-art algorithms}

\begin{table*}[tb]
    \caption{Learning conditions for SAC and DSAC-T:
    the methods combined in the proposed algorithm use the default hyperparameters given in the references presented, unless otherwise noted.
    }
    \label{tab:param_sac}
    \centering
    {\scriptsize
    \begin{tabular}{lc}
        \hline\hline
        \#Hidden layers of fully connected networks & $2$
        \\
        \#Neurons for each hidden layer & $100$
        \\
        Activation function & Squish \citep{barron2021squareplus,kobayashi2023design}
        \\
        & + RMSNorm \citep{zhang2019root}
        \\
        Estimation model of each value function & Two independent models \citep{haarnoja2018soft}
        \\
        Distribution model of policy & Modified PERT distribution
        \\
        \hline
        Discount factor $\gamma$ & $0.99$
        \\
        Optimizer & AdaTerm \citep{ilboudo2023adaterm}
        \\
        The way of updating target networks & CAT-soft update \citep{kobayashi2024consolidated}
        \\
        \hline
        Buffer size $|\mathcal{D}|$ & $102400$
        \\
        Batch size & $32$
        \\
        \#Replayed data & $|\mathcal{D}|/8$
        \\
        Hyperparameters of PER \citep{schaul2016prioritized} $(\alpha^{\mathrm{PER}}, \beta^{\mathrm{PER}})$ & $(1, 0.5)$
        \\
        \hline\hline
    \end{tabular}
    }
\end{table*}

\begin{figure*}[tb]
    \begin{subfigure}[b]{0.32\linewidth}
        \centering
        \includegraphics[keepaspectratio=true,width=\linewidth]{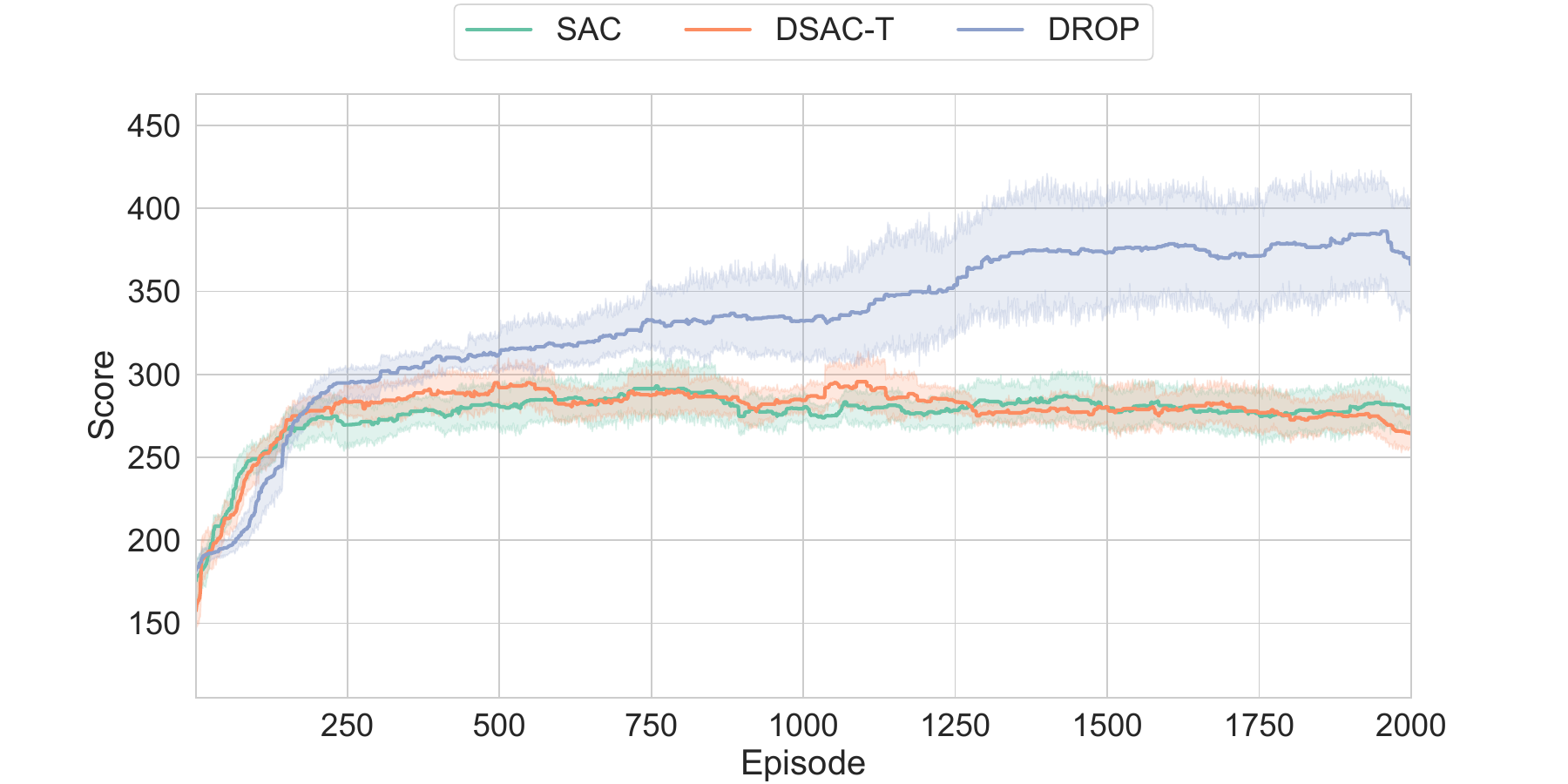}
        \subcaption{TwoPoles}
        \label{fig:score_TwoPoles}
    \end{subfigure}
    \begin{subfigure}[b]{0.32\linewidth}
        \centering
        \includegraphics[keepaspectratio=true,width=\linewidth]{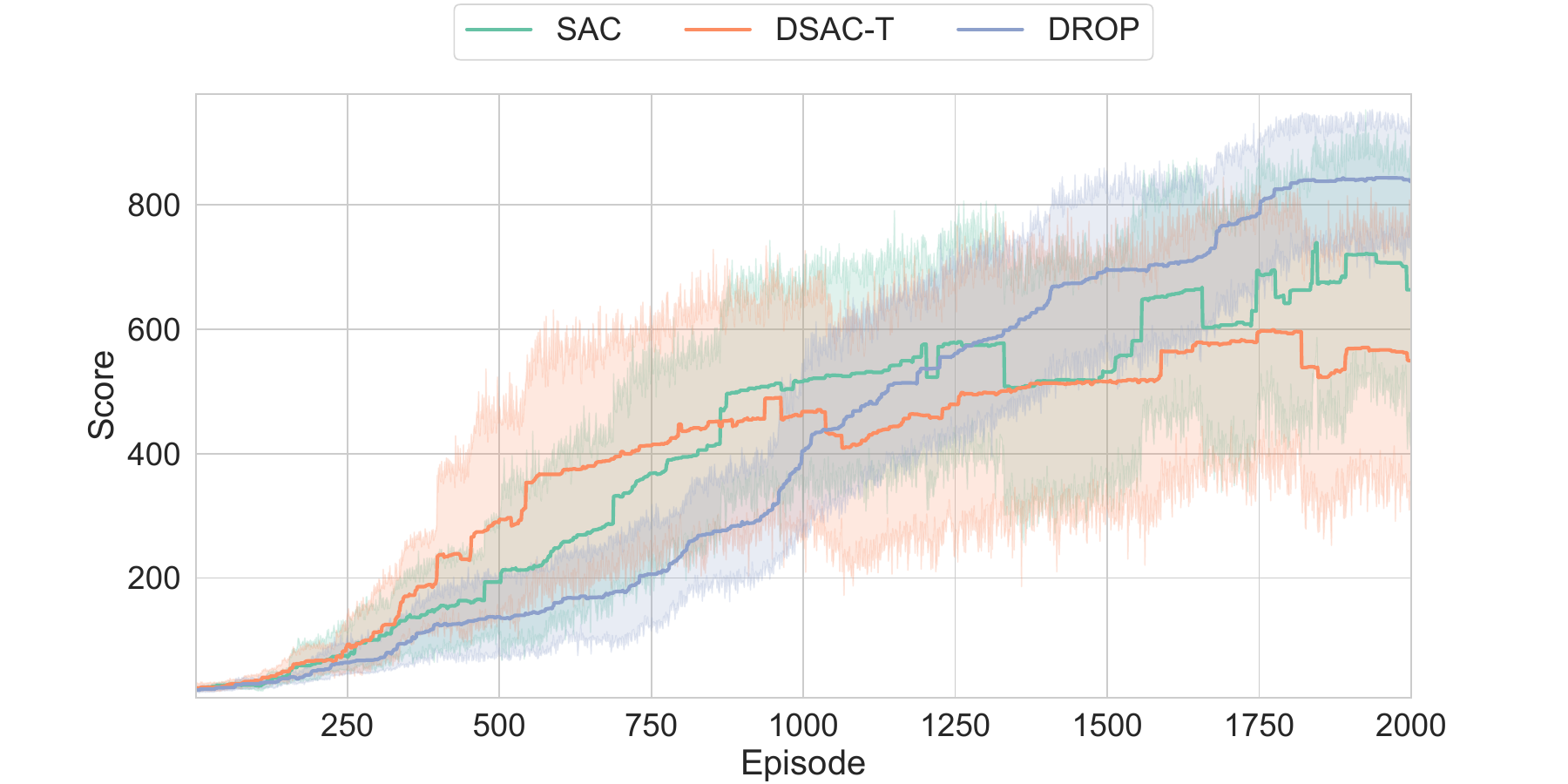}
        \subcaption{HopperStand}
        \label{fig:score_HopperStand}
    \end{subfigure}
    \begin{subfigure}[b]{0.32\linewidth}
        \centering
        \includegraphics[keepaspectratio=true,width=\linewidth]{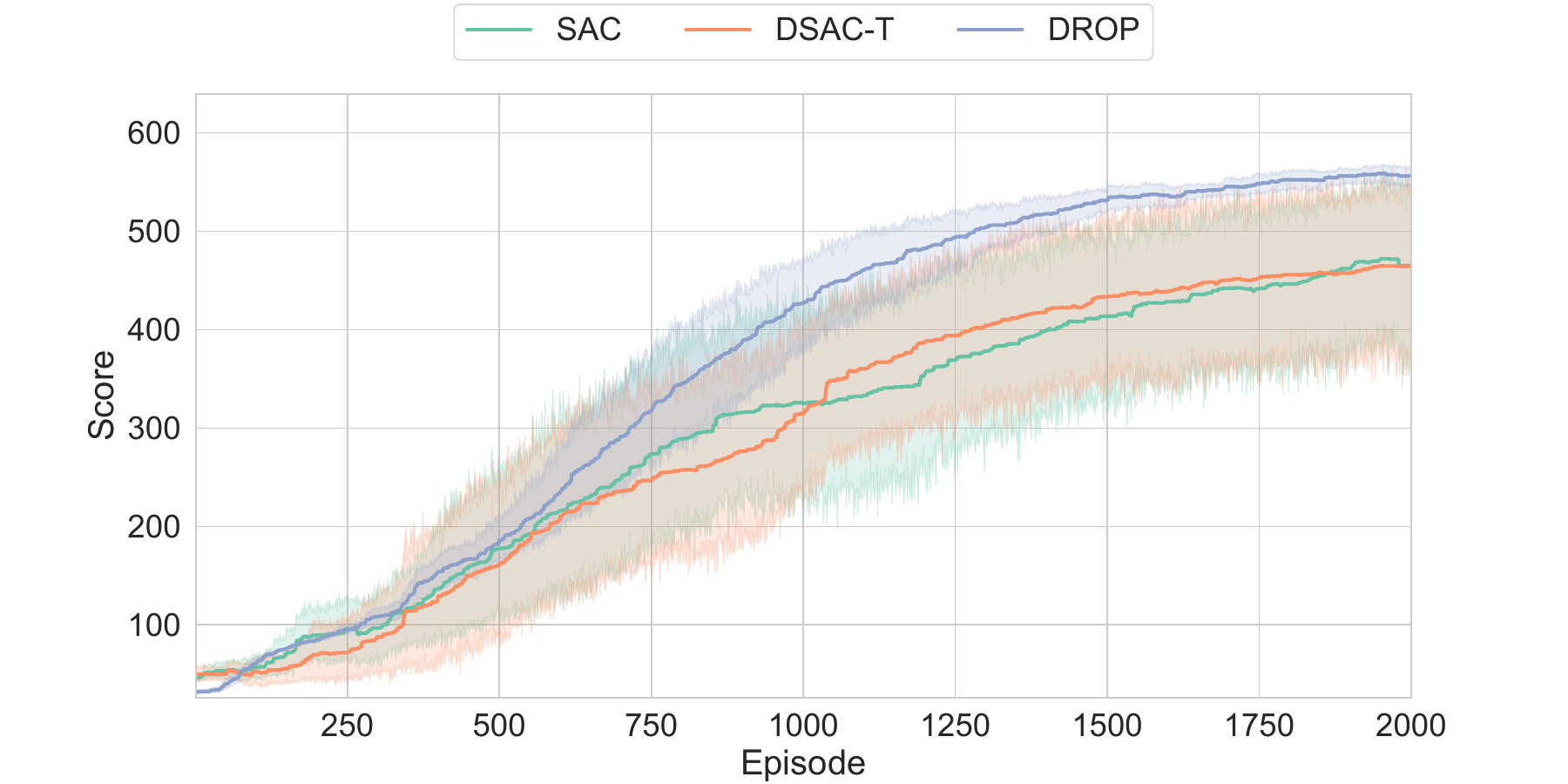}
        \subcaption{WalkerRun}
        \label{fig:score_WalkerRun}
    \end{subfigure}
    \caption{Comparisons to the state-of-the-art algorithms:
    the comparisions, SAC and DSAC-T, trapped into local optima;
    the proposed DROP succeeded in acquiring all the tasks stably.
    }
    \label{fig:comparison}
\end{figure*}

As shown above, the theoretically-grounded DROP has better generalization performance than the previous heuristic model.
This section further demonstrates the value of DROP by comparing it to other state-of-the-art algorithms.
The comparison is made between soft actor-critic (SAC) \citep{haarnoja2018soft}, the leading state-of-the-art algorithm in non-distributional RL, and its extension to distributional RL, so-called DSAC-T \citep{duan2025distributional}.
The core implementation and their hyperparameters are based on the original papers, but to ensure a fair comparison, the peripheral implementations that can be shared with DROP (such as network architecture, optimizer, experience replay, and so on) were made identical to DROP wherever possible (see Table~\ref{tab:param_sac}).
Note that, since it has been pointed out that transforming the policy model using a tanh function that forcibly restricts the output to the range $[-1, 1]$ prevents the analytical derivation of the mean of the probability distribution \citep{chen2024rethinking}, the modified PERT distribution with the same domain is employed as an alternative.
In addition, ERC \citep{kobayashi2024revisiting} is not needed in SAC and DSAC-T since they are off-policy algorithms, which can adopt experience replay as they are.
L2C2 \citep{kobayashi2022l2c2} is also removed because maximizing policy entropy in SAC and DSAC-T serves a similar role.

By conducting a different benchmark from the above experiments, DROP can also be demonstrated that it has the capability to learn a wider variety of tasks.
Specifically, the tasks for verification are the following three, which are prepared in \textit{dm\_control} \citep{tunyasuvunakool2020dm_control}.
\begin{itemize}
    \item \textit{TwoPoles}:
    Let a passive bilink connected to a cart swing up and stand.
    While it is densely rewarded, it requires sufficient exploration to exit the local solution.
    \item \textit{HopperStand}:
    Let a planar one-legged robot stand, starting in a random posture.
    With the sparse reward, the value function has a non-Gaussian distribution shape.
    \item \textit{WalkerRun}:
    Let a planar biped robot stand and run at 8~m/s.
    If it falls over even once, the reward easily decreases.
\end{itemize}
Note that these API is converted to the Gymnasium API by Shimmy.
The original time step of \textit{TwoPoles} was 0.01~s, which was too short to accomplish the task, so it was changed to 0.02~s, increasing the time of one episode to 20~s accordingly.
Although the number of tasks is smaller than in the above experiments, as described later, these alone clearly demonstrate the difference in learning performance between DROP and baselines.

The learning curves of 12 random seeds are statistically depicted in Fig.~\ref{fig:comparison}.
As can be seen, there was no noticeable difference in performance between the baseline SAC and DSAC-T.
This is because DSAC-T assumes a Gaussian distribution of the value function, which could not adequately represent the true shape of the distribution, and thus could not provide the full benefit of it.
In all the tasks, SAC and DSAC-T fell into a local solution, although the learning speed was fast.
Both SAC and DSAC-T attempt to improve policies based on relatively pessimistic values, and thus tend to converge on conservative policies, even though policy entropy facilitates exploration stochastically.

In contrast, only DROP was able to reach a global solution stably.
Since DROP adopts the median TD error from an ensemble of values for policy improvement, it is likely that both optimistic and pessimistic improvements are balanced.
Furthermore, it is possible that in addition to the random exploration by stochastic policies, the change in the characteristics of the improvement over time worked well as a directed exploration \citep{wilson2014humans}.
In any case, DROP was found to have superior learning performance compared to recent RL algorithms.

\section{Conclusion and future work}

This paper investigated a novel theoretically-grounded model of RL derived on the basis of the control as inference, which replaces the heuristic model with asymmetric learning rates to explain the real behaviors of dopamine neurons reported in the recent studies \citep{dabney2020distributional,muller2024distributional}.
First, the derivation of optimistic model in the previous study \citep{kobayashi2022optimistic} was reviewed, and then, a pessimistic model was theoretically derived by considering the inversion of the definition of optimality.
Then, the degree of optimism and pessimism was made adjustable with a unified bounded parameter, and both optimism and pessimism were obtained in a balanced manner by regularly spacing multiple parameters based on an ensemble method.
While the heuristic model sometimes failed in learning, the proposed method, DROP, could succeed in stable learning with appropriately small TD errors.

Thus, the heuristic model used to explain the real data was not useful as a RL algorithm, but DROP was useful as it, partly because it was theoretically grounded.
Conversely, using DROP to explain real data may enable fitting with higher accuracy than the heuristic model.
Therefore, it would be worthwhile to use DROP as a model to explain the real data.

On the other hand, there is still room for improvement in DROP as a RL algorithm.
Specifically, DROP can make effective use of the distributional value function, which is omitted in this paper for simplicity.
Depending on how it is used, DROP should be effective in various ways, such as ensuring safety through worst-case optimization, efficient exploration prioritizing the best-case scenario, and so on.
By comparing with the latest algorithms, such as those introduced in the related work, one can expect to improve DROP and make it more practical as an RL algorithm.
Alternatively, it would be interesting to see if the consideration and introduction of biologically-plausible meta-objective, which optimizes the statistic $\mathcal{M}$, can also explain the policy improvement in animals.

\subsection*{Acknowledgments}

This work was supported by JSPS KAKENHI, Development and validation of a unified theory of prediction and action, Grant Number JP24H02176.

\appendix
\section{Estimated value distribution}
\label{app:distribution}

\begin{figure}[tb]
    \centering
    \includegraphics[keepaspectratio=true,width=0.96\linewidth]{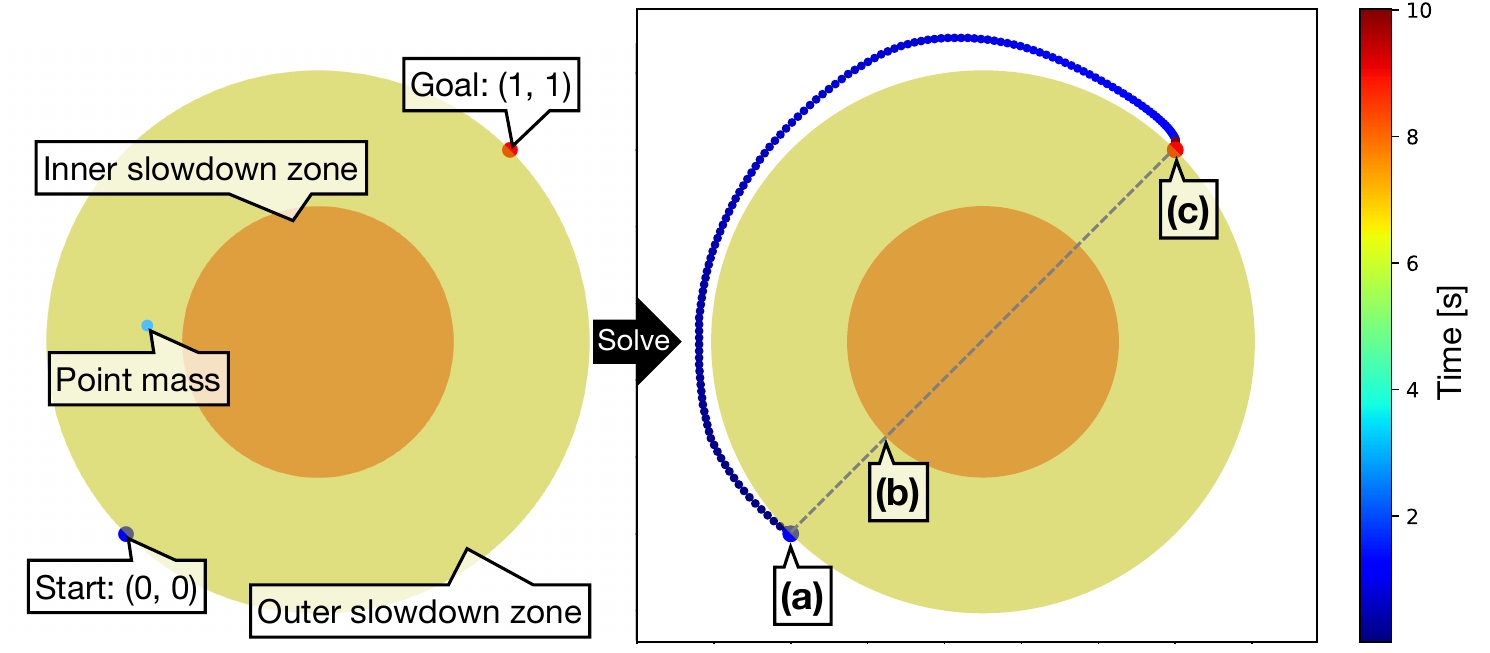}
    \caption{Toy problem setup and example of optimal trajectory:
    in the yellow outer zone, negative rewards occur when speed exceeds the threshold of $0.5$;
    in the orange inner zone, exceeding the threshold ends the episode with a large negative reward;
    the optimal solution is to bypass these zones and head toward the goal.
    }
    \label{fig:toy_env}
\end{figure}

\begin{figure}[tb]
    \centering
    \includegraphics[keepaspectratio=true,width=0.96\linewidth]{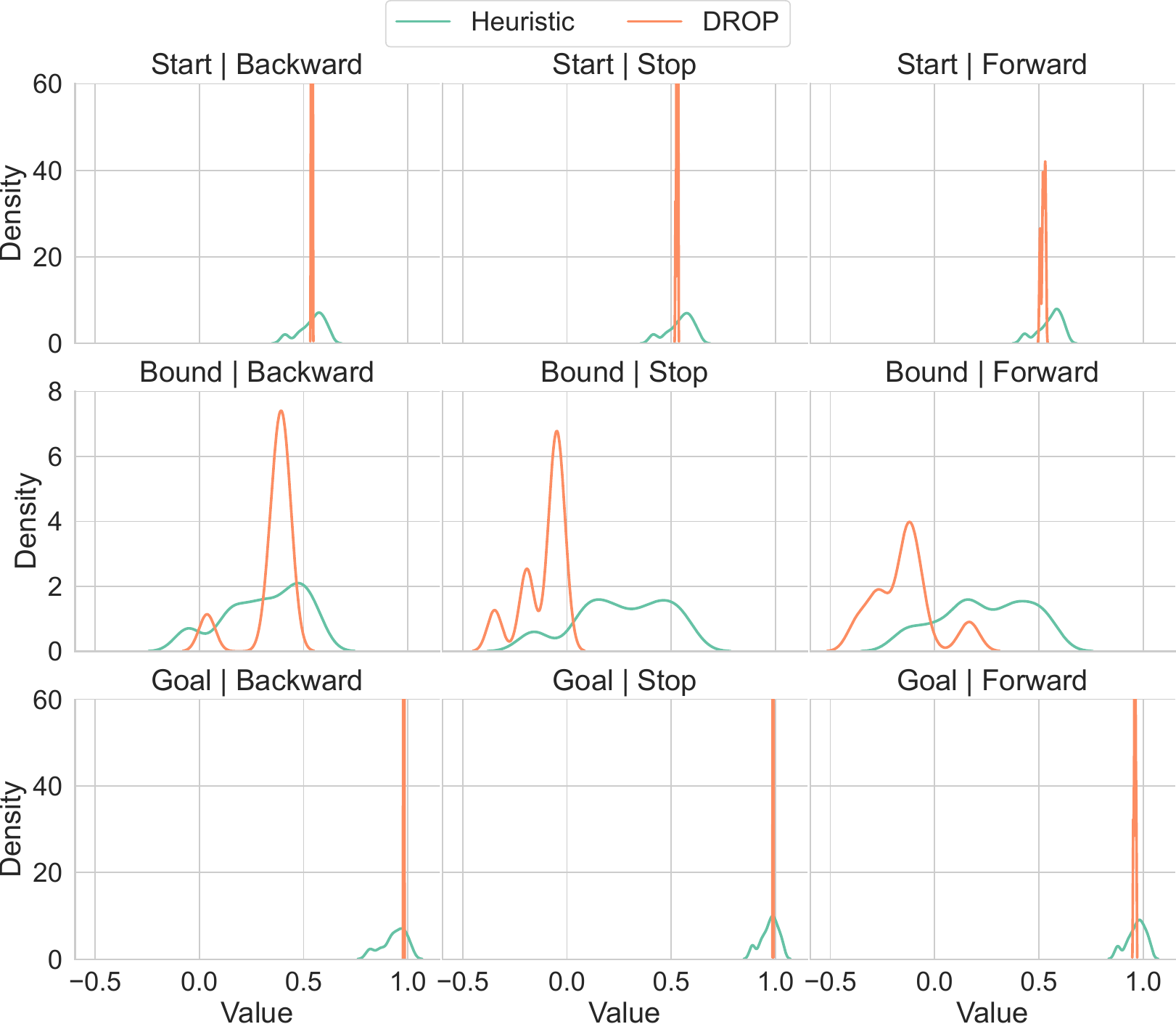}
    \caption{Estimated value distributions:
    DROP accurately estimated the (multimodal) value distributions for all the states;
    \textit{Heuristic} captured qualitative features of the distributions, but estimation errors caused the distributions to broaden.
    }
    \label{fig:toy_value}
\end{figure}

To verify that the proposed DROP with the architecture depicted in Fig.~\ref{fig:value_dist} can approximate the value distribution, an exemplification is conducted using the following toy problem.
Specifically, the toy problem involves the movement of a point on a two-dimensional plane toward a goal position (see Fig.~\ref{fig:toy_env} left).
The action is a two-dimensional target velocity.
The next velocity is the weighted average of this and the current one (the weight is randomly selected with uniform probability).
The position is then updated using the Euler method.
Therefore, the state consists of a two-dimensional position and velocity.

Between the starting position and the goal lies slowdown zones where speed penalties apply.
In the outer zone, negative rewards proportional to excess speed are given.
In the inner zone, exceeding the speed limit is treated as failure, ending the episode with a large negative reward.
Therefore, the optimal solution involves traveling along a path that bypasses the slowdown zones.
Maintaining high speed within the slowdown zones causes a probabilistic speed penalty regardless of action, resulting in a multimodal value distribution.

An example of solving this toy problem using the proposed DROP is illustrated in Fig.~\ref{fig:toy_env} right.
As expected, the agent avoided the slowdown zones and reached the goal.
This was also the case with \textit{Heuristic} used in Section~\ref{subsubsec:comparison}.
After obtaining the optimal policy, learning was continued to achieve sufficient convergence of the value function.

To visualize the estimated value distribution, three positions shown in Fig.~\ref{fig:toy_env} right (\textit{Start}, \textit{Bound}, and \textit{Goal}) and three velocities (\textit{Backward} with $(-2, -2)$, \textit{Stop} with $(0, 0)$, and \textit{Forward} with $(2, 2)$) are chosen.
The value distributions for these combinations are depicted in Fig.~\ref{fig:toy_value}.

First, let's focus on the results of DROP.
At \textit{Start} and \textit{Goal}, the estimated value distributions exhibited high density at all velocities.
This indicates that the optimal trajectory was drawn with the optimal policy, effectively counteracting the influence of probabilistic behavior, and that the corresponding value was obtained appropriately.
However, with \textit{Forward} at \textit{Start} (upper right), which attempts to enter the outer zone, the impact of losses incurred until escaping from there caused bimodality.
Meanwhile, \textit{Bound} clearly estimated multimodal value distributions.
First, \textit{Backward} fundamentally assigned high value since there is no risk of entering the inner zone, but it also had another mode near zero value, reflecting the risk of failing to maintain a velocity that avoids penalties in the outer zone.
In \textit{Stop}, the overall value was lower due to the significant loss incurred until escaping the outer zone.
It appropriately captured both the risk of incurring penalties in the outer zone (the middle mode) and the risk of failing by entering the inner zone (the left mode).
Furthermore, \textit{Forward} correctly estimated the heightened risk of entering the inner zone and failing.
The right mode had higher value than that of \textit{Stop} because it represented the potential to efficiently escape the slowdown zones by utilizing part of the initial velocity.

On the other hand, focusing on \textit{Heuristic}, the estimated value distribution was qualitatively similar to DROP in all states but was clearly broad.
Particularly, the distributions were also broad at \textit{Goal}, exceeding the theoretical maximum return in this toy problem.
In other words, although the value function should converge sufficiently during training, estimation errors still remained.
This is likely due in part to the discontinuity inherent in the learning rule of \textit{Heuristic}, which destabilized the learning process.

Thus, despite locally updating each unit of the value function in the implementation used in this paper, it was able to accurately estimate the value distribution.
This is thought to be due to the effective influence of implicitly incorporating information from other units by sharing the networks up to the output layer.
That is, while the local updates can be interpreted as introducing the sampling bias from the value distribution, it might be mitigated by the fact that each estimated value function is influenced by others.

\section{Performance profiles}
\label{app:performance}

\begin{figure*}[tb]
    \begin{subfigure}[b]{0.48\linewidth}
        \centering
        \includegraphics[keepaspectratio=true,width=\linewidth]{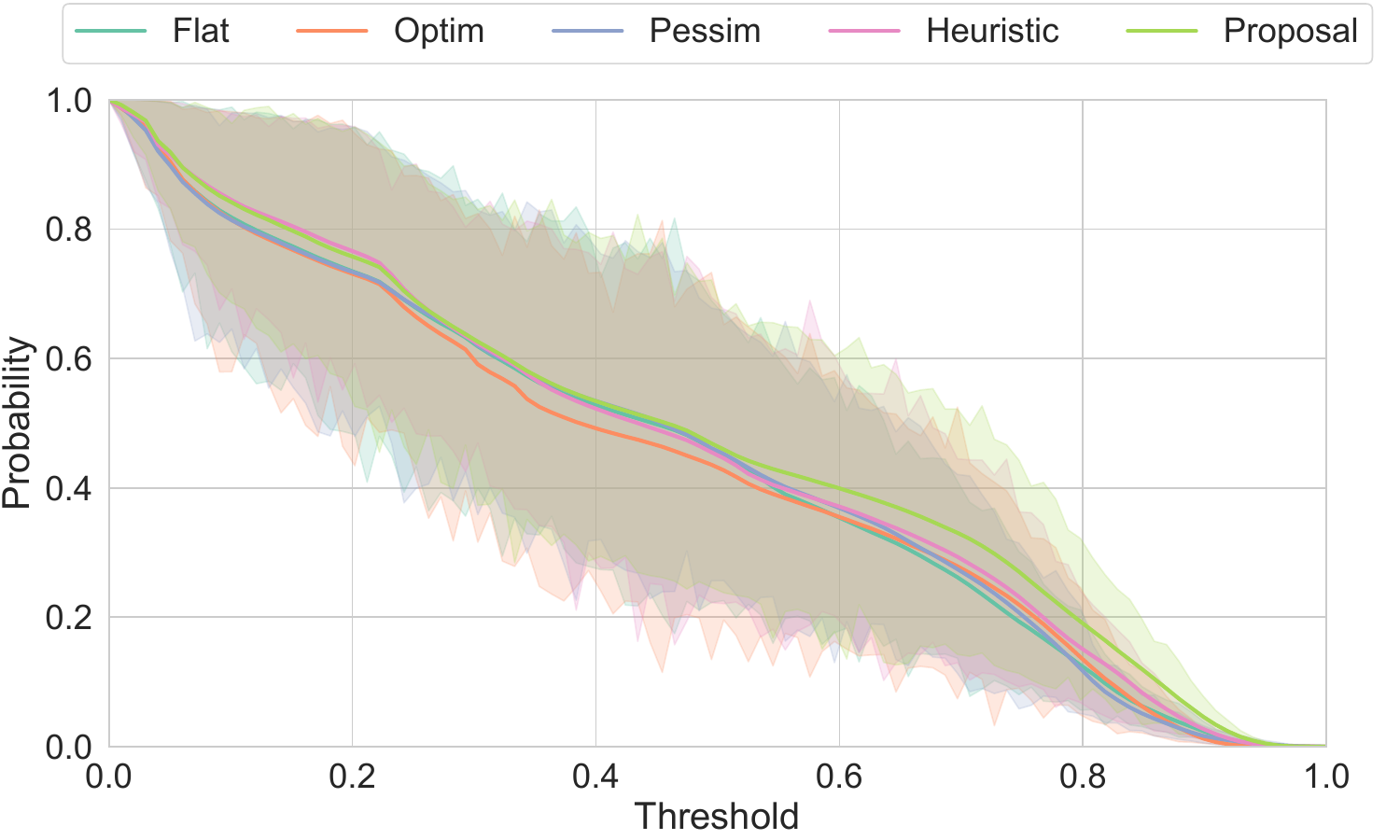}
        \subcaption{Profile in the latter half of learning}
        \label{fig:profile_learn}
    \end{subfigure}
    \begin{subfigure}[b]{0.48\linewidth}
        \centering
        \includegraphics[keepaspectratio=true,width=\linewidth]{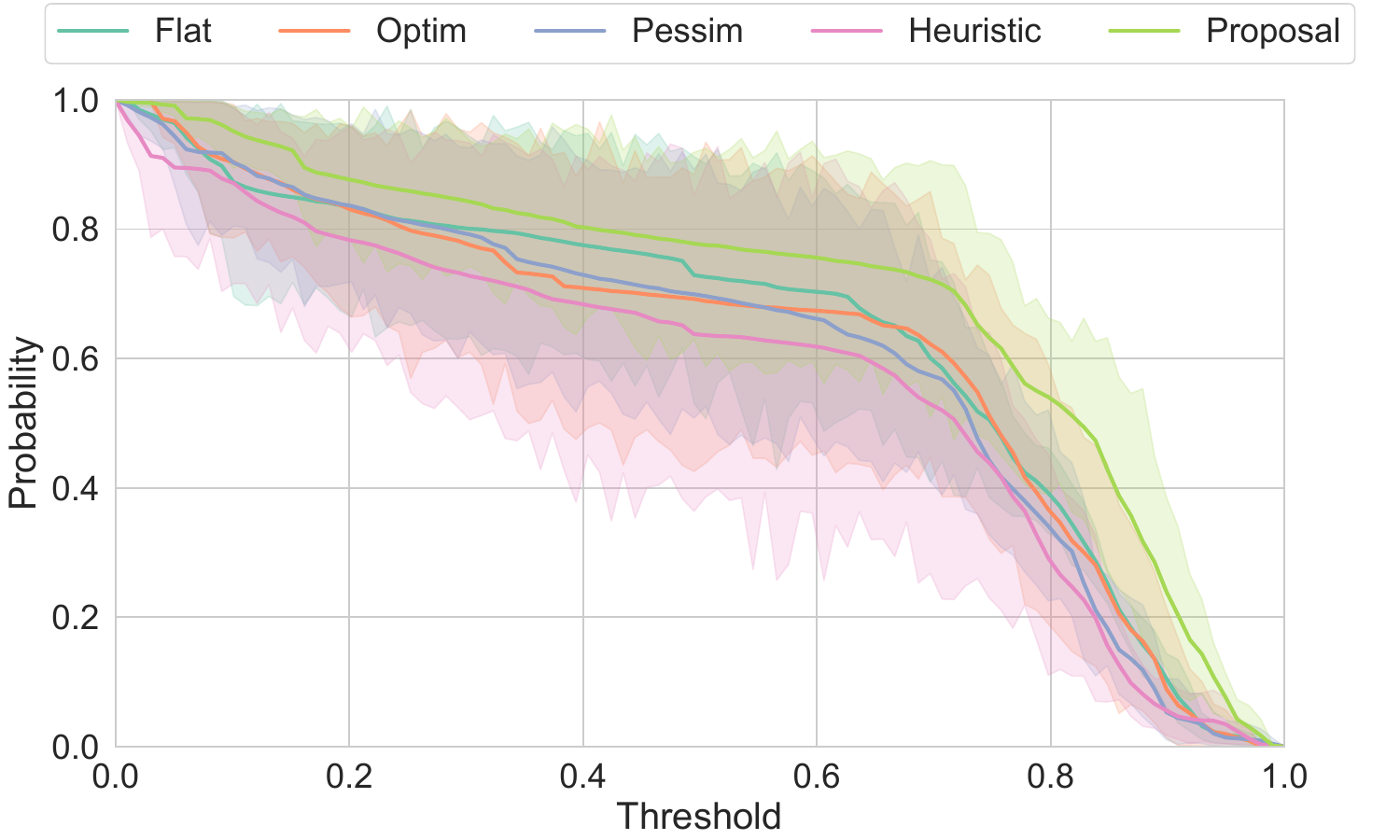}
        \subcaption{Profile in post-learning tests}
        \label{fig:profile_test}
    \end{subfigure}
    \caption{Performance profiles for the experiments in Section~\ref{subsubsec:comparison}:
    the proposed DROP achieved the faster, higher-quality, and more robust control performances than the baselines including the heuristic model.
    }
    \label{fig:profile}
\end{figure*}

A deeper analysis is conducted for the experimental results presented in Section~\ref{subsubsec:comparison}.
To this end, performance profiles suggested in \citep{agarwal2021deep} are employed.
Specifically, all scores for each task are normalized to $[0, 1]$ using the maximum and minimum scores, and the probability that the score exceeds a given threshold $\tau \in [0, 1]$ is evaluated.
Here, two types of returns are considered as scores:
i) the return for each episode in the latter half of learning to understand the convergence rate and speed;
ii) the return when testing the tasks with the post-learning policies to understand learning quality and robustness.

These results are plotted in Fig.~\ref{fig:profile}.
Note that while this curve generally slopes downward, a slower descent indicates better performance.
The four baseline methods performed at a similar level.
In contrast, only the proposed DROP method consistently achieved higher scores than the others.
While Fig.~\ref{fig:result_comp} focused on control performance for each task after learning, this comprehensive evaluation (including the learning process) also concludes that the proposed method is superior.

\bibliographystyle{apalike}
\bibliography{biblio}

\end{document}